%% file: preprint.tex
\documentclass[11pt]{article}

\PassOptionsToPackage{table}{xcolor}

\usepackage[preprint]{acl}

\usepackage{times}
\usepackage{latexsym}

\usepackage[T1]{fontenc}

\usepackage[utf8]{inputenc}

\usepackage{microtype}

\usepackage{inconsolata}

\usepackage{graphicx}

\usepackage{algorithm}
\usepackage{algorithmic}
\usepackage{mathtools}
\usepackage{tcolorbox}

\definecolor{myblue}{HTML}{F0F0F0}

\usepackage{booktabs}
\usepackage{multirow}
\usepackage{tikz}
\usetikzlibrary{arrows.meta,positioning}

\title{Repairing LLM-Generated Specifications with Deterministic Diagnosis}

\author{
 \textbf{Giovanni Vaccarino\textsuperscript{1,2,3}},
 \textbf{Emanuele La Malfa\textsuperscript{1,4}},
 \textbf{Dehai Min\textsuperscript{2}},
 \textbf{Hanrong Zhang\textsuperscript{2}},
 \textbf{Pier Luca Lanzi\textsuperscript{3}},
\\
 \textbf{Lu Cheng\textsuperscript{5}},
 \textbf{Michael J. Wooldridge\textsuperscript{1}},
 \textbf{Philip S. Yu\textsuperscript{2}},
 \textbf{Sara Bernardini\textsuperscript{1}}
\\
\\
 \textsuperscript{1}Department of Computer Science, University of Oxford, Oxford, UK\\
 \textsuperscript{2}Department of Computer Science, University of Illinois Chicago, Chicago, IL, USA\\
 \textsuperscript{3}Dipartimento di Elettronica, Informazione e Bioingegneria, Politecnico di Milano, Milan, Italy\\
 \textsuperscript{4}Institute for Decentralized AI (IDAI)\\
 \textsuperscript{5}Department of Computer Science and Engineering, The Pennsylvania State University, University Park, PA, USA
\\
 \small{\textbf{Correspondence:} \href{mailto:giovanni.vaccarino@mail.polimi.it}{giovanni.vaccarino@mail.polimi.it}}
\\
 \small{\raisebox{-0.2em}{\includegraphics[height=1.1em]{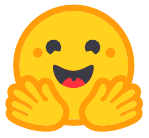}}\,\href{https://huggingface.co/datasets/gianniiii/planetarium-floor-tile-corrected}{Data release}}
\\
}

\begin{document}

\maketitle

\input{sections/abstract}

\input{sections/introduction}

\input{sections/related_work}

\input{sections/method}

\input{tables/main_results}

\input{sections/experiments}

\input{sections/results}

\input{sections/discussion}

\input{sections/conclusion}

\bibliography{custom}

\clearpage

\appendix
\input{sections/appendix}

\end{document}

%% file: sections/abstract.tex
\begin{abstract}
Large language models (LLMs) are increasingly used as formalizers: given a reasoning problem expressed in natural language, they produce a formal specification that a symbolic solver can then solve, with guarantees relative to that specification. Many specifications are runnable yet semantically wrong, encoding incorrect constraints, domains, objectives, predicates, or goals, and solver-error feedback dries up once execution succeeds. 
Purely LLM-based diagnosis, lacking a formal ground truth, often fails to spot these errors, and its ungrounded critiques cause regressions in $11.8$\% of initially correct specifications in our experiments. We propose DRAGO, a repair framework that replaces this ungrounded critique with deterministic, formalism-specific diagnosis, and retains a rewrite only when the diagnosed error weight strictly decreases under a reference-free acceptance predicate. Across five benchmarks spanning logical reasoning (SMT-LIB2), constraint programming (MiniZinc), and planning (PDDL), DRAGO cuts regressions by nearly an order of magnitude, to $1.3$\%, while improving answer accuracy by $12.9$ percentage points over zero-shot formalization and $8.7$ percentage points over LLM-diagnostic repair, averaged across benchmarks. We also uncover two rendering defects in Planetarium that cause $50$\% ($3{,}086$ instances) of its floor-tile split to be inconsistent with the reference formalizations, and release a corrected split.
 
\end{abstract}

%% file: sections/introduction.tex
\section{Introduction}

In the paradigm of Large Language Models (LLMs) as formalizers, an LLM turns a natural language problem specification into a formal specification~\cite{ liu2023llmpempoweringlargelanguage, olausson-etal-2023-linc, cai2025galagloballlmagents}. Such a paradigm separates the linguistic interpretation from the combinatorial machinery that a solver provides, and delegates searching, proving, or validating an (optimal) solution to a symbolic backend~\cite{pan-etal-2023-logic, ye2023satlm, ahmaditeshnizi2024optimus, hao2025planning, huang-zhang-2025-limit}. 

A fundamental challenge beyond generating executable code with LLMs is that of ensuring that the resulting artifact preserves the intended semantics of the original problem. In fact, semantic failures often go unnoticed, as solvers cannot provide guidance on resolving them~\cite{ ni2026draftandpruneimprovingreliabilityautoformalization,peran2026largelanguagemodelsoptimizers}. 
A generated artifact may parse and run while still encoding an incorrect quantifier scope, missing a constraint, mistyping a domain, or specifying the wrong goal. In this sense, solver-based feedback loops are weak: while parse, type, runtime, and status failures provide a rich repair signal~\cite{pan-etal-2023-logic, ahmaditeshnizi2024optimus}, that signal becomes sparse precisely when an incorrect artifact becomes executable, leaving the LLM with no actionable feedback for improving.
Recent mitigations include LLM-based diagnosis, which supplies denser feedback by asking a \emph{reviewer} (e.g., another LLM~\cite{pmlr-v235-kambhampati24a}) to critique the generated formalization. However, when no reference formalization or ground-truth answer is available, the reviewer must infer the errors from the same problem statement and artifact that produced the mistake. This can result in missed errors, false alarms, and repairs that corrupt previously correct artifacts~\cite{huang2024large, stechly2025on, tyen-etal-2024-llms}; Figure~\ref{fig:fol-transitions} in Sec. 5 illustrates this failure mode directly. The failure is not merely that self-review misses errors; it is that its false alarms actively corrupt correct artifacts, often breaking more than they repair. Since a reference formalization is typically unavailable at test time, repair requires reliable, reference-free feedback grounded in observable evidence.

We propose \textbf{D}esign, \textbf{R}epair \textbf{A}nd \textbf{GO} (\textbf{DRAGO}), a formalization-repair framework centered on automatic reference-free diagnosis. DRAGO first generates a candidate formalization, then applies deterministic, formalism-specific checks to the candidate and its observable execution behavior. These checks produce structured findings, which are converted into repair hints and provided to a repair model for artifact revision. The diagnostic rules are deliberately formalism-specific, but the repair pattern is shared and replaces ungrounded LLM-based diagnosis with automatic reference-free feedback.
Beyond diagnosis, DRAGO retains a proposed rewrite only if it satisfies an evidence-gated acceptance predicate that requires the diagnosed error weight to decrease strictly.  The underlying design principle is that repairs should be \emph{aimed} and \emph{accepted} based on observable deterministic evidence rather than by the model's own ungrounded judgment. This principle provides a common repair interface that decouples the roles of grounded diagnosis and evidence-gated acceptance~\cite{lian2026reloopstructuredmodelingbehavioral,ao2026optirepairclosedloopdiagnosisrepair,ni2026draftandpruneimprovingreliabilityautoformalization}.

We instantiate DRAGO on three domains: logical reasoning, constraint programming, and automated planning. These domains are operationalized through SMT-LIB2 artifacts~\cite{BST10} for first-order logic, MiniZinc models~\cite{1771668.1771709} for constraint programming, and PDDL specifications~\cite{mcdermott_pddl_1998} for planning. We fix one diagnostic configuration per formalism and apply it consistently across all benchmarks and models evaluated within that formalism. Across these settings, DRAGO improves answer accuracy by an average of $12.9$ percentage points over zero-shot generation and by $8.7$ percentage points over LLM-diagnostic repair. Crucially, the two methods differ most in what they break: LLM-diagnostic repair corrupts more initially correct formalizations than it fixes, while DRAGO's regression rate is nearly an order of magnitude lower ($1.3\%$ vs.\ $11.8\%$ of initially correct artifacts). We also highlight a common evaluation issue in LLM formalization: benchmarks themselves are subject to errors and ambiguities. In Planetarium, we uncover two rendering defects affecting 50\% of the floor-tile test split and release a corrected version. Our main contributions are:
\begin{itemize}
    \item We identify runnable-but-wrong artifacts as a recurrent failure mode in LLM formalization across three formalisms and show that LLM-diagnostic repair is unreliable precisely in this silent region, producing both missed repairs and right-to-wrong regressions.
    \item We introduce \textbf{DRAGO}, a repair interface shared across formalisms that replaces LLM-based diagnosis with automatic, reference-free diagnosis grounded in deterministic checks and solver probes, and combines it with evidence-gated acceptance.
    \item Across five benchmarks and four models, DRAGO cuts right-to-wrong regressions by nearly an order of magnitude relative to LLM-diagnostic repair ($1.3$\% vs.\ $11.8$\% of initially correct artifacts), while improving answer accuracy by $12.9$ percentage points over zero-shot formalization and $8.7$ percentage points over the LLM-diagnostic baseline.
    \item Highlighting the broader challenge of reliable evaluation in LLM formalization, we identify two previously undocumented rendering defects that make 50\% ($3{,}086$ instances) of Planetarium's floor-tile test split inconsistent with its references. We patch all affected instances and release the corrected split as open source.
\end{itemize}

Below, we present DRAGO’s formal framework, evaluate it across three formalization domains, and discuss its empirical behavior, broader implications, and limitations.

%% file: sections/related_work.tex
\section{Related Work}

\paragraph{Solver-backed LLM formalization.} 
LLMs used as front ends for symbolic systems appear in logical reasoning, where Logic-LM, SatLM, and LINC connect language models to symbolic solvers~\cite{olausson-etal-2023-linc, pan-etal-2023-logic, ye2023satlm}; in planning, LLM+P and recent planning-formalization work translate natural language into PDDL-like or programmatic representations~\cite{liu2023llmpempoweringlargelanguage, hao2025planning, huang-zhang-2025-limit}; in optimization and constraint modeling, systems and benchmarks study LLM generation of executable models for mathematical or constraint solvers~\cite{ahmaditeshnizi2024optimus, cai2025galagloballlmagents, michailidis2026dcpbenchopenevaluatingllmsconstraint}.  Works that directly employ chain-of-thought and self-consistency~\cite{wei2022chain, wang2023selfconsistency}, remain limited on multi-step planning and constraint reasoning tasks~\cite{valmeekam2023planbenchextensiblebenchmarkevaluating, lin2025zebralogic}.
Solver-backed formalization is often more attractive than asking an LLM to carry out the full solution search due to the guarantees the former provides.  However, this paradigm shifts the central challenge from solving the problem to translating it faithfully: the generated formal artifact must preserve the intended semantics of the natural-language specification.

\paragraph{Repair from execution, validation, and symbolic diagnosis.} 
Logic-LM feeds solver error messages into self-refinement~\cite{pan-etal-2023-logic}, OptiMUS executes and debugs generated optimization models~\cite{ahmaditeshnizi2024optimus}, while self-debugging applies the same loop to general code generation~\cite{chen2024teaching}.  
Planning pipelines additionally consult validators, such as VAL~\cite{10.1109/ICTAI.2004.120}, and PDDL-construction systems iterate against validator or planner feedback~\cite{guan2023leveraging, gestrin2024towards, peran2026largelanguagemodelsoptimizers}. Similar principles have also been applied to automated program repair, symbolic infeasibility diagnosis, and model-based diagnosis~\cite{10.1109/TSE.2011.104, ISSTA-2015-QiLAR, 10.1007/s10817-007-9084-z, f2798bd6728d4bf0b6a2557dee66a226, REITER198757}. Unlike approaches that localize faults against a test suite~\cite{orvalho2025counterexample} or plans of known feasibility~\cite{lin2025told}, DRAGO diagnoses and accepts repairs from the problem statement and the artifact alone, through one interface shared across formalisms.

\paragraph{From self-critique to grounded semantic verification.} 
LLM-based diagnosis tools, such as Self-Refine and Reflexion, iterate on model-generated feedback~\cite{madaan2023selfrefine, shinn2023reflexion}, with agentic modeling systems that apply the pattern to formalization~\cite{ahmaditeshnizi2024optimus, cai2025galagloballlmagents}.  
Without a reference, however, their unreliability is well documented: intrinsic self-correction degrades reasoning accuracy~\cite{huang2024large}, self-verification fails on planning and constraint-style tasks~\cite{stechly2025on}, self-repair for code often does not repay its cost~\cite{olausson2024is}, and models fail at \emph{locating} errors while repairing well once the error is localized for them~\cite{kamoi-etal-2024-llms,tyen-etal-2024-llms}. 
An emerging line of work seeks to incorporate external evidence and verification~\cite{gou2024critic, pmlr-v235-kambhampati24a}: recent works pursue this goal through sampling diversity and pruning~\cite{ni2026draftandpruneimprovingreliabilityautoformalization}, parallel synthesis with LLM-written semantic checkers~\cite{song2026cpsyncmultiagentzeroshotconstraint}, behavioral probing~\cite{lian2026reloopstructuredmodelingbehavioral}, agentic closed-loop diagnosis of optimization models~\cite{ao2026optirepairclosedloopdiagnosisrepair}, and learned alignment judges~\cite{lu2025formalalign}.

%% file: sections/method.tex
\section{Methodology}
\label{sec:meth}

Let $\Sigma^*$ be the Kleene closure of a finite vocabulary $\Sigma$, where all finite-length natural language strings live.
Let $x \in \mathcal{X} \subseteq \Sigma^*$ be a natural-language problem instance that describes, for example, a planning or an optimization problem, and
$\mathcal{L} \subset \Sigma^*$ be a solver-facing formal language.  A formalization of $x$ is an artifact $p \in \mathcal{L}$ executed by a backend $S$, namely
\[
  S:\mathcal{L}\rightarrow\mathcal{E} \subset \Sigma^*,
  \qquad e=S(p),
\]
where $e$ is the observable execution trace exposed by the solver stack.  The
trace may contain parse status, runtime errors, solver status, proof
outcomes, objective values, produced assignments, or planner/evaluator
metadata.  
A parametrized model (e.g., an LLM), which manipulates task $x$, acts as a \emph{translator} between natural language and a formal language $\mathcal{L}$ and its solver $S$. We denote this (stochastic) "translation" and its (sampled) output as $p \sim G(\cdot | x)$, where $G$ is the generation model.

While LLMs can generate simple formal language specifications in a single iteration~\cite{hao2025planningrigorgeneralpurposezeroshot} (i.e., in \emph{zero-shot}), they struggle with complex planning and reasoning problems. To address this limitation, several techniques generate a candidate solution and iteratively refine it through sequential calls to the model. Below, we describe the most common techniques and their limitations.

\subsection{Existing Repair Feedback and Its Limitations}
\label{sec:repair-feedback}

For a natural language specification $x$, we define its $t$-th repair step as a function of the current artifact $p_t$, trace
$e_t=S(p_t)$, and a history $h_t=\{x, (p_j,e_j,z_j)_{j<t}\}$, where $z_j$ is the feedback supplied to the repair model.  
In this sense, a generic repair policy has the form
\begin{equation}\label{eq:repair-feedback}
    p_{t+1} \sim R(\cdot\mid x,p_t,e_t,z_t,h_t),
\end{equation}
where $p_{t+1}$ is a proposed rewrite. In the equation above and throughout the rest of the paper, we denote with $R$ a repair or a generation that comes from a model (e.g., an LLM) that \emph{possibly} differs from $G$. Different repair systems mainly differ in
how they construct $z_t$. 

In the \emph{zero-shot} setting, the repair policy described above is never invoked, and the solution to the problem is $p \sim G(\cdot | x)$.
On the other hand, multiple rounds of repair feedback impose an upper bound on $t < T$ in Equation~\ref{eq:repair-feedback}. 
This neurosymbolic approach has two major limitations, which we articulate in the following paragraphs.

\paragraph{Execution Feedback.} 
When a solver successfully executes a program $p_t$, any potential signal that may indicate inconsistencies between the original specification $x$ and $p_t$ vanishes; a syntactically valid $p_t$ may still diverge from the semantic intent of $x$ (e.g., omitted constraints or incorrect objectives), leaving uninformative execution feedback.
Formally, for a deterministic feedback solver:
\[
  z_t = \phi(e_t) = 
  \begin{cases} 
    \texttt{error trace}, & e_t \in \mathcal{E}_{err} \\
    \emptyset, & e_t \in \mathcal{E}_{sil}
  \end{cases}
\]
where $\mathcal{E} = \mathcal{E}_{err} \cup \mathcal{E}_{sil}$ is a partition of the trace space into execution failures and silent successes: $\mathcal{E}_{err}$ refers to programs that return compilation errors, runtime exceptions, timeouts, resource-limit violations, etc. (\texttt{error trace}), with $\mathcal{E}_{sil}$ complementing $\mathcal{E}_{err}$ with programs that do not return any of the above mentioned failures.
While highly reliable in $\mathcal{E}_{err}$, the signal vanishes in the \emph{silent region} $\mathcal{E}_{sil}$.

\paragraph{LLM-based Diagnosis.} 
To overcome the sparsity of $\mathcal{E}_{sil}$, self-review methods~\cite{pmlr-v235-kambhampati24a} generate dense, stochastic feedback via an LLM critique policy $R$, $z_t \sim R(\cdot \mid x, p_t)$.
Because $R$ evaluates alignment without a ground-truth reference $p^* \in \mathcal{L}$, it must infer mismatches directly from $x$ and $p_t$. This yields a noisy evaluation channel prone to false negatives (missing the errors of an incorrect $p_t$) and false positives (hallucinating flaws and rewriting an already-correct artifact).

\subsection{DRAGO: Deterministic Diagnosis and Evidence-Gated Repair}
\label{sec:drago}
Algorithm~\ref{alg:drago} summarizes DRAGO's repair loop. DRAGO implements a series of interventions on the candidate program $p_t$ to mitigate the execution and self-review feedback issues discussed above. 
In doing so, DRAGO keeps the LLM as the reviser but switches the feedback source for $z_t$ to a deterministic diagnostic operator
\[
  D:\mathcal{X}\times\mathcal{L}\times\mathcal{E}\rightarrow\mathcal{Z},
  \qquad z_t=D(x,p_t,e_t),
\]
implemented without additional LLM calls or oracle access. The operator interface is common to all approaches (CSP, planning, first-order logic, etc.), but is converted into each formalism to produce solver-specific rewrites.

\paragraph{Reference-free Diagnostic Classes.}
DRAGO's diagnostic output is a finite set of findings, defined as:
\[
  D(x,p,e)=\{d_k\}_{k=1}^{K}, \qquad
  d_k=(c_k,a_k,\eta_k,\rho_k).
\]
Here, $c_k$ is an error category, $a_k$ is the affected artifact fragment or semantic feature, $\eta_k$ is the observable evidence supporting the diagnosis, and $\rho_k$ is a local repair instruction. The evidence $\eta_k$ must be derivable solely from $(x,p,e)$ and additional reference-free solver probes; it cannot depend on benchmark labels or reference artifacts. Every check DRAGO performs falls into one of three classes. (i) \emph{Contract checks} verify artifact-internal invariants without requiring a solver call. They ensure that the required output structure aligns with the declared vocabulary, types, and arities of the provided signature or data schema. (ii) \emph{Solver probes} perform additional reference-free executions that turn an observed symptom into a localized cause: premise-consistency and entailment probes in FOL, deletion probes that localize conflicting requirement blocks on unsatisfiable MiniZinc instances, in the spirit of deletion-based MUS extraction~\cite{10.1007/s10817-007-9084-z}, and delete-relaxation reachability proving goal atoms unachievable in PDDL. (iii) \emph{Specification-alignment checks} compare structures extracted from the problem statement by fixed rules against the artifact, e.g., stated object counts with declared objects in PDDL. Purely lexical cues are advisory by design: they never trigger a rewrite on their own.

Ultimately, this interface provides the repair model with a narrower, more precise edit distribution than ungrounded self-review. Instead of asking the model to blindly rediscover whether anything is wrong, DRAGO tasks it with repairing a concrete, localized discrepancy, $q_t \sim R(\cdot\mid x,p_t,e_t,z_t,h_t)$, where, for $z_t=\emptyset$, no semantic repair is performed.
Appendix~\ref{app:diagnostic-rules} details the full instantiation of each class.

\paragraph{Evidence-Gated Acceptance.}

An LLM rewrite can introduce new errors, so DRAGO never retains a rewrite on its own authority. The repair model proposes a candidate $q_t$, and the system keeps it only if the acceptance predicate holds:
\[
  p_{t+1} =
  \begin{cases}
    q_t, & A(x,p_t,q_t,e_t,\tilde e_t,z_t)=1,\\
    p_t, & \text{otherwise}.
  \end{cases}
\]

Let $w(D(x,p,e))\ge 0$ denote the total weight of the findings returned by the diagnostic operator; $w$ counts the blocking findings, with multiplicity when one finding covers several violations, and advisory findings contribute zero. The predicate $A$ holds only when three conditions are met: (i) \emph{well-formedness}: the candidate parses, executes, and preserves the required vocabulary and output structure; (ii) \emph{diagnostic progress}: re-running the diagnostic operator strictly reduces the error weight, $w(D(x,q_t,\tilde e_t)) < w(D(x,p_t,e_t))$; and (iii) \emph{conservatism}: rewrites that regress the solver status, or relax the artifact while improving the objective, are rejected, and edits that change the answer-determining status are accepted only with independent reference-free corroboration. The acceptance predicate is not a correctness oracle, as it cannot prove that $q_t$ matches the ground truth semantics, and it never consults the benchmark answer. Further, because every accepted rewrite strictly decreases a discrete nonnegative diagnostic potential, the sequence of accepted artifacts cannot cycle. The loop stops when $w(D(x,p_t,e_t))=0$, a proposal is rejected, or the repair budget is exhausted. Appendix~\ref{app:diagnostic-rules} describes the concrete diagnostic checks, acceptance conditions, and implementation defaults for each formalism.  

\begin{algorithm}[tb]
\caption{DRAGO formalization repair}
\label{alg:drago}
\small
\begin{algorithmic}[1]
\REQUIRE instance $x$, language $\mathcal{L}$, executor $S$, generator
$G$, repair model $R$, diagnostic operator $D$, acceptance check
$A$, budget $T$.
\STATE $p \sim G(\cdot \mid x,\mathcal{L})$;\quad $h_0 \leftarrow [\,]$
\FOR{$t=0,\ldots,T-1$}
  \STATE $e_t \leftarrow S(p)$;\quad $z_t \leftarrow D(x,p,e_t)$
  \STATE \textbf{if} $w(z_t)=0$ \textbf{then break}
  \STATE $q_t \sim R(\cdot \mid x,p,e_t,z_t,h_t)$;\quad $\tilde{e}_t \leftarrow S(q_t)$
  \STATE \textbf{if} $A(x,p,q_t,e_t,\tilde{e}_t,z_t)=0$ \textbf{then break}
  \STATE $h_{t+1}\leftarrow h_t\cup\{(p,e_t,z_t)\}$;\quad $p\leftarrow q_t$
\ENDFOR
\STATE \textbf{return} $p$
\end{algorithmic}
\end{algorithm}

%% file: tables/main_results.tex
\begin{table*}[!t]
\centering
\small
\setlength{\tabcolsep}{2.5pt}
\renewcommand{\arraystretch}{1.05}
\newcommand{\sd}[1]{{\scriptsize$\pm#1$}}
\newcommand{\imp}[1]{{\scriptsize\,$(#1)$}}
\resizebox{\textwidth}{!}{%
\begin{tabular}{@{}llcccccccccccc@{}}
\toprule
& & \multicolumn{3}{c}{Haiku 4.5} & \multicolumn{3}{c}{Sonnet 4.6}
& \multicolumn{3}{c}{Qwen3-Coder 30B} & \multicolumn{3}{c}{Gemma 4 31B} \\
\cmidrule(lr){3-5}\cmidrule(lr){6-8}\cmidrule(lr){9-11}\cmidrule(lr){12-14}
Domain & Benchmark & Zero & LLM-d & DRAGO & Zero & LLM-d & DRAGO
& Zero & LLM-d & DRAGO & Zero & LLM-d & DRAGO \\
\midrule
\multirow{4}{*}{FOL} & FOLIO
& $75.0$ & $81.5$ & \cellcolor{myblue}{$\mathbf{85.5}$}
& $85.0$ & $81.2$ & \cellcolor{myblue}{$\mathbf{90.1}$}
& $60.9$ & $63.7$ & \cellcolor{myblue}{$\mathbf{64.8}$}
& $85.0$ & $85.8$ & \cellcolor{myblue}{$\mathbf{87.6}$} \\[-2pt]
&
& \sd{0.7} & \sd{1.9} & \sd{0.8}\imp{+4.0}
& \sd{0.2} & \sd{2.1} & \sd{1.3}\imp{+5.1}
& \sd{0.5} & \sd{1.8} & \sd{0.3}\imp{+1.1}
& \sd{0.9} & \sd{1.1} & \sd{0.4}\imp{+1.8} \\
\addlinespace
& ProverQA
& $77.1$ & $74.4$ & \cellcolor{myblue}{$\mathbf{89.4}$}
& $77.2$ & $74.1$ & \cellcolor{myblue}{$\mathbf{90.3}$}
& $60.8$ & $59.8$ & \cellcolor{myblue}{$\mathbf{68.8}$}
& $75.1$ & $74.2$ & \cellcolor{myblue}{$\mathbf{83.7}$} \\[-2pt]
&
& \sd{0.3} & \sd{0.8} & \sd{0.5}\imp{+12.3}
& \sd{0.4} & \sd{1.0} & \sd{0.2}\imp{+13.1}
& \sd{0.4} & \sd{1.1} & \sd{0.6}\imp{+8.0}
& \sd{0.1} & \sd{0.3} & \sd{0.1}\imp{+8.6} \\
\midrule
\multirow{4}{*}{CSP} & Text2Zinc
& $17.5$ & $29.6$ & \cellcolor{myblue}{$\mathbf{42.2}$}
& $40.4$ & $52.0$ & \cellcolor{myblue}{$\mathbf{57.1}$}
& $12.0$ & $\mathbf{22.2}$ & \cellcolor{myblue}{$19.8$}
& $20.4$ & $38.7$ & \cellcolor{myblue}{$\mathbf{56.5}$} \\[-2pt]
&
& \sd{0.8} & \sd{1.5} & \sd{1.0}\imp{+12.6}
& \sd{1.0} & \sd{1.2} & \sd{1.2}\imp{+5.1}
& \sd{0.9} & \sd{1.4} & \sd{1.8}\imp{-2.4}
& \sd{2.5} & \sd{2.0} & \sd{1.2}\imp{+17.8} \\
\addlinespace
& CP-Bench
& $35.6$ & $55.9$ & \cellcolor{myblue}{$\mathbf{62.9}$}
& $61.6$ & $69.8$ & \cellcolor{myblue}{$\mathbf{79.7}$}
& $20.3$ & $25.1$ & \cellcolor{myblue}{$\mathbf{42.9}$}
& $43.9$ & $59.3$ & \cellcolor{myblue}{$\mathbf{70.9}$} \\[-2pt]
&
& \sd{1.8} & \sd{2.4} & \sd{1.8}\imp{+7.0}
& \sd{1.3} & \sd{1.9} & \sd{1.3}\imp{+9.9}
& \sd{1.2} & \sd{2.3} & \sd{2.7}\imp{+17.8}
& \sd{2.7} & \sd{3.3} & \sd{0.8}\imp{+11.6} \\
\midrule
\multirow{2}{*}{PDDL} & Planetarium
& $58.0$ & $58.7$ & \cellcolor{myblue}{$\mathbf{60.7}$}
& $79.7$ & $70.3$ & \cellcolor{myblue}{$\mathbf{80.3}$}
& $15.2$ & $8.0$ &  \cellcolor{myblue}{$\mathbf{16.3}$}
& $61.2$ & $60.9$ & \cellcolor{myblue}{$\mathbf{69.4}$} \\[-2pt]
&
& \sd{0.3} & \sd{0.3} & \sd{0.3}\imp{+2.0}
& \sd{0.3} & \sd{0.3} & \sd{0.3}\imp{+0.6}
& \sd{0.5} & \sd{1.3} & \sd{1.2}\imp{+1.1}
& \sd{0.7} & \sd{1.5} & \sd{1.0}\imp{+8.2} \\
\bottomrule
\end{tabular}%
}
\caption{Main answer accuracy results, including our approach, DRAGO (gray background). LLM-d denotes the LLM-diagnostic repair baseline. Cells report
mean accuracy percentages with the sample standard deviation across
independent runs. Planetarium is evaluated as released on $300$ sampled
problems, of which $58$ are inconsistent with their references, capping
accuracy at $80.7\%$ (Section~\ref{sec:measurement-validity}). \textbf{Bold} marks the best method for
each benchmark within each model. The parenthesized $(+x)$ beneath each DRAGO mean is its gain over the best baseline.}
\label{tab:main-results}
\end{table*}

%% file: sections/experiments.tex
\section{Experimental Setting}

\paragraph{Domains, Benchmarks, and Models.}

We evaluate DRAGO across three formalization domains: constraint programming (CSP), first-order logic (FOL), and planning (PDDL).  These domains are operationalized, respectively, through MiniZinc models~\cite{1771668.1771709} on Text2Zinc~\cite{singirikonda2025text2zinccrossdomaindatasetmodeling} and CP-Bench~\cite{michailidis2026dcpbenchopenevaluatingllmsconstraint}, SMT-LIB2 artifacts~\cite{BST10} on FOLIO~\cite{han2024folionaturallanguagereasoning} and ProverQA-hard~\cite{qi2025large}, and PDDL specifications~\cite{mcdermott_pddl_1998} on Planetarium~\cite{zuo-etal-2025-planetarium}. Planetarium is evaluated as released, with its original problem texts and references, on a fixed sample of $300$ test problems ($100$ per domain); $58$ of the sampled floor-tile problems are inconsistent with their references (Section~\ref{sec:measurement-validity}), so achievable accuracy is capped at $80.7\%$. Full benchmark details are given in Appendix~\ref{app:benchmark-details}.
We evaluate two models from the Anthropic family, Sonnet~4.6~\cite{anthropic2026claudesonnet46systemcard} and Haiku~4.5~\cite{anthropic2025claudehaiku45systemcard}, together with two open-source models with different architectures: Qwen3-Coder 30B~\cite{yang2025qwen3technicalreport}, a mixture-of-experts (MoE) model, and Gemma 4 31B~\cite{gemmateam2026gemma4technicalreport}, a dense model. Additional information about the models is reported in Appendix~\ref{app:implementation-details}.

\paragraph{Metrics.}
The primary metric is instance-level accuracy under each benchmark's scorer. For FOL and CSP, this is answer-level correctness (the entailment label and the objective value or feasibility, respectively), which a formalization can achieve without matching the intended semantics exactly (we address that in the Limitations). For PDDL, the scorer instead measures semantic equivalence between the generated problem and the reference. We report mean accuracy and the sample standard deviation across independent runs. We also use exact paired McNemar tests~\cite{McNemar1947} on item-level correctness, adjusted with the Holm--Bonferroni correction~\cite{29def780-e117-38f0-8afb-edf384af3fad}.  Tests are reported at the domain level, merging the benchmarks of a formalism (Appendix~\ref{app:statistical-analysis}).

\paragraph{Baselines.}
We compare DRAGO with two baselines. In \emph{zero-shot} generation, the generator $G_\theta$ produces a single formalization $p_0$, which we execute with the domain backend $S$ and score directly. \emph{LLM-diagnostic repair} extends this setup by using the same LLM as diagnostic and repair model: non-runnable artifacts receive ordinary solver-error feedback as $z_t$, while for runnable artifacts one self-review call diagnoses possible modeling errors and a second call proposes the rewrite $q_t$. DRAGO keeps the same generation and repair model and differs only in the diagnosis and the acceptance check (Section~\ref{sec:meth}).

\paragraph{Implementation Details.}
All current Anthropic calls use provider-default sampling settings. Open-source models are run using the vLLM engine, on a single node of NVIDIA 4xA100, serving with recommended model sampling parameters.  Both the LLM-diagnosis baseline and DRAGO use a one-round budget. Each formalism uses a single DRAGO configuration throughout the experiments across the four evaluated models and the benchmarks of that domain. Each configuration was developed on a small development set disjoint from every evaluated instance; the sources are detailed in Appendix~\ref{app:diagnostic-rules}. Exact model identifiers, inference parameters, software versions, timeouts, and repair budgets are listed in Appendix~\ref{app:implementation-details}, and prompt templates in Appendix~\ref{app:prompt-templates}.

%% file: sections/results.tex
\section{Experimental Results}
\label{sec:results}
\subsection{Main Results}
\label{sec:main-results}
Table~\ref{tab:main-results} shows that DRAGO is the strongest overall repair method: it improves accuracy over zero-shot generation for every model on every benchmark and achieves the best performance in $19$ of $20$ model-benchmark comparisons.  The gains over zero-shot are statistically significant (Holm-corrected McNemar tests) for FOL and CSP on both hosted models and for PDDL on Haiku (all $p<.0001$; Appendix Table~\ref{tab:domain-mcnemar}, with paired gains and $95\%$ intervals in Table~\ref{tab:paired-effect-sizes}). DRAGO’s advantage over LLM-diagnostic repair depends on whether ungrounded self-review is beneficial or detrimental. Where it degrades the artifacts (FOL for both hosted models, PDDL for Sonnet), DRAGO's advantage is large and significant. Even on CSP, where self-review also helps, DRAGO still achieves a higher mean on every benchmark.

FOL and CSP largely benefit from DRAGO, which leverages a formalization with enough structure to diagnose the semantics gap.  This is clearest on ProverQA-hard, where both hosted models gain about $13$ points, and on CSP, where DRAGO adds up to $27$ points for the hosted models and up to $36$ points for Gemma.  The LLM-diagnostic baseline is far less stable, falling below zero-shot on ProverQA-hard for both hosted models and on FOLIO for Sonnet.  DRAGO's one loss is on Text2Zinc with Qwen, where it improves over zero-shot but trails LLM-diagnostic repair by $2.4$ points, possibly because many of the zero-shot artifacts are too far from a correct model for targeted repair to help.

Of the three domains that we test, PDDL is the most challenging to marginally improve.
The reasons include a previously unknown benchmark defect that caps achievable accuracy at approximately $80.7\%$ (see Section~\ref{sec:measurement-validity}), alongside well-known issues with PDDL benchmarks for LLMs (such as their performance ``saturation''~\cite{lamalfa2026endtoendpddlplanninghardcoded}). 
Despite the achievable performance cap, DRAGO's gains are consistent, ranging from 0.6 percentage points for Sonnet, whose zero-shot accuracy already sits at the ceiling ($239$ of the $242$ answerable problems), to $+8.2$ percentage points for Gemma with almost no regressions. By contrast, unconditional self-review is destructive, costing Sonnet $9.4$ points and Qwen $7.2$ (per-subdomain results in Appendix Table~\ref{tab:pddl-subdomains}). 

\begin{figure}[t]
\centering
\includegraphics[width=0.85\columnwidth]{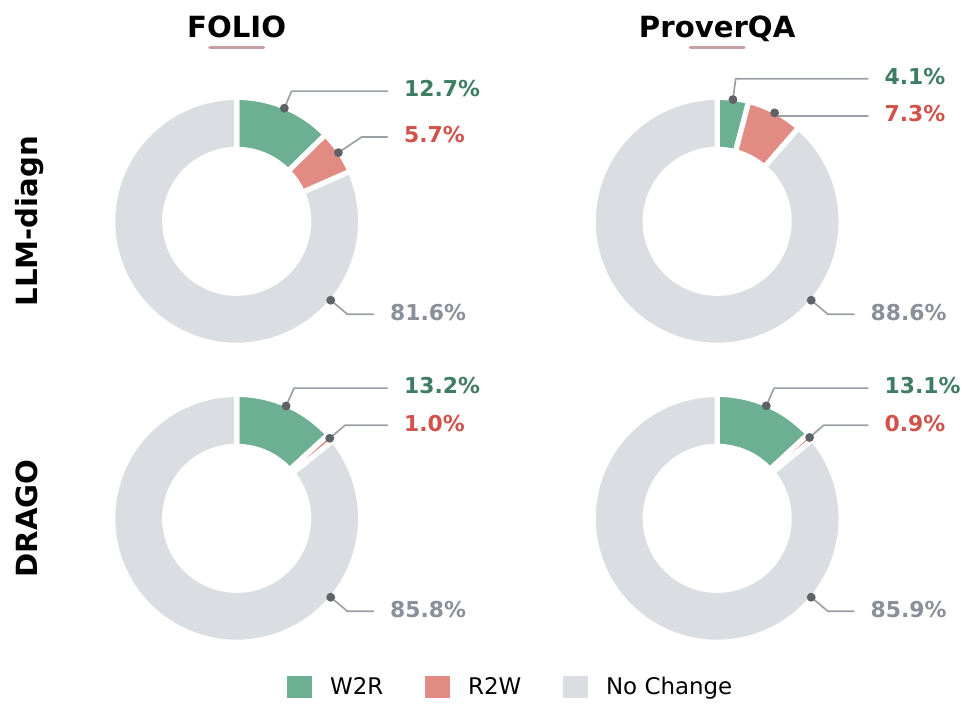}
\caption{Correctness transitions of the repair loop on the FOL benchmarks
(Haiku) for the LLM-diagnostic baseline and DRAGO on FOLIO and
ProverQA-hard. Slices are shares of all instances, pooled over runs.}
\label{fig:fol-transitions}
\end{figure}

\subsection{Do Repairs Fix More Than They Break?}
\label{sec:transitions}

Accuracy alone hides \emph{why} a repair pipeline helps or hurts: a method can raise accuracy by recovering wrong programs (wrong$\rightarrow$right, W2R) while silently corrupting correct ones (right$\rightarrow$wrong, R2W). Figure~\ref{fig:fol-transitions} decomposes every instance into W2R, R2W, and no-change (both W2W and R2R) for Haiku across the FOL benchmarks. The decisive difference is in regressions: as a share of all instances, the baseline breaks $5.7\%$ on FOLIO and $7.3\%$ on ProverQA-hard, where it breaks more than it fixes, while DRAGO holds regressions at about $1\%$. As a share of initially correct artifacts, the regression rates are $7.7\%$ and $9.4\%$ for the baseline against $1.3\%$ and $1.2\%$ for DRAGO, respectively. In terms of recovery, DRAGO repairs $13.2\%$ and $13.1\%$ of instances on the two benchmarks, respectively, whereas the baseline's recovery is $12.7\%$ on FOLIO and $4.1\%$ on ProverQA-hard. The remaining models and benchmarks show the same pattern (Appendix~\ref{app:additional-results}), most starkly on PDDL: pooled over all $2{,}400$ paired rows, DRAGO moves $104$ instances from wrong to right against $12$ regressions, whereas self-review breaks more artifacts than it fixes ($254$ against $159$). Across the runs, LLM-diagnostic repair breaks $11.8\%$ of initially correct artifacts ($1{,}006$ of $8{,}559$), more than it fixes, whereas DRAGO breaks $1.3\%$ ($110$ of $8{,}528$), nearly an order of magnitude fewer.

\begin{table}[b]
\centering
\small
\setlength{\tabcolsep}{5.0pt}
\begin{tabular}{@{}lcccc@{}}
\toprule
& Zero & Sample-3 & Oracle-3 & DRAGO \\
\midrule
FOL & $78.6$ & $80.0$ & $81.3$ & $\mathbf{88.8}$ \\
CSP & $39.2$ & $44.1$ & $46.0$ & $\mathbf{59.9}$ \\
PDDL & $68.9$ & $69.7$ & $71.9$ & $\mathbf{70.2}$ \\
\bottomrule
\end{tabular}
\caption{Sample-and-select inference against DRAGO, aggregated over the
hosted-model configurations of each formalism. \textbf{Bold} marks the best
implementable method.  Selection rules and per-configuration results are in
Appendix~\ref{app:ablations}, Table~\ref{tab:sampling-full}.}
\label{tab:sampling-summary}
\end{table}

\subsection{Diagnosis or Acceptance Gate?}
\label{sec:ablations}

\begin{figure}[t]
\centering
\includegraphics[width=\columnwidth]{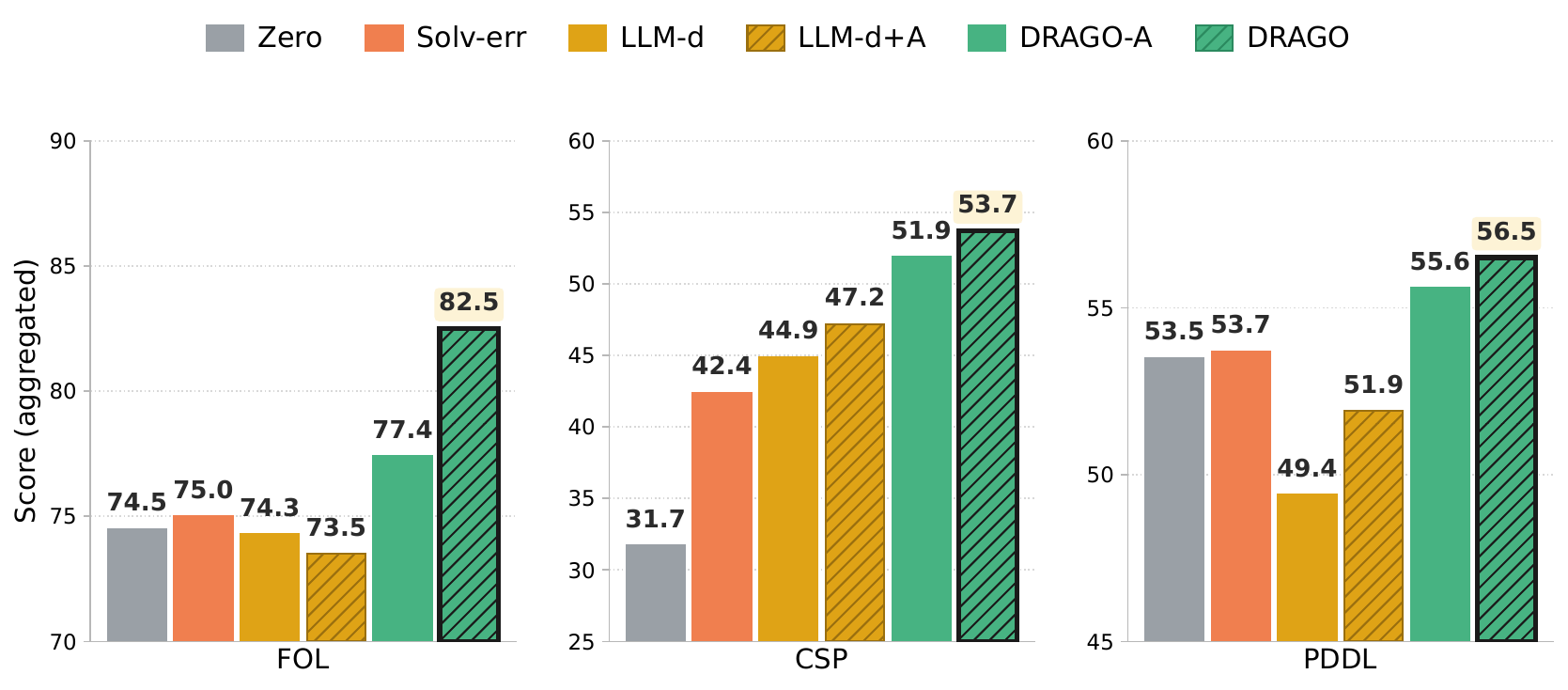}
\caption{Component ablations aggregated
over the model-benchmark configurations of each formalism. Per-configuration results and exact arm definitions in
Appendix~\ref{app:ablations}. Best arm per
formalism in \textbf{bold}.}
\label{tab:ablation-summary}
\end{figure}

This section isolates the contribution of the DRAGO deterministic diagnosis from the acceptance predicate through three ablation variants. We instantiate \emph{DRAGO$-$A}, which preserves the diagnosis while accepting every rewrite, \emph{LLM-d$+$A}, which augments LLM self-review with DRAGO's acceptance predicate, and \emph{solver-error-only}, which restricts repairs to artifacts that fail during execution. Figure~\ref{tab:ablation-summary} summarizes the results across formalisms.
Diagnosis has the larger effect: holding acceptance fixed, DRAGO outperforms LLM-d$+$A in $16$ of $20$ configurations, by $+7.1$ points on average and by about $+18$ on ProverQA-hard for both hosted models. The gate matters, as a rewrite can flip the final label: removing it (DRAGO$-$A) costs $5.1$ points in FOL, $1.8$ in CSP, and $0.9$ in PDDL. Conversely, gating (LLM-d$+$A) shifts self-review by only $+1.1$ points on average, and in every configuration where self-review loses at least a point to zero-shot, the gated variant stays below zero-shot (Appendix~\ref{app:ablations}).  Finally, solver-error-only repair confirms the silent-failure taxonomy presented in Table~\ref{tab:silent-failure-taxonomy}: it matches zero-shot performance on FOL and PDDL, where nearly all incorrect artifacts execute cleanly, and improves accuracy by 10.7 percentage points on CSP, the one domain where execution failures still dominate, on top of which DRAGO's semantic layer adds an additional $11.3$ percentage-point improvement. The ablations separate the two effects: grounded diagnosis is responsible for repairing incorrect artifacts, and the acceptance gate is responsible for preserving correct ones.

%% file: sections/discussion.tex
\section{Discussion}
\label{sec:discussion}

\subsection{State-of-the-Art Comparison}
\label{sec:alternatives}

To further validate DRAGO, we compare it against two additional baselines: sample-and-select inference, which spends its budget on diversity rather than repair, and state-of-the-art domain-specific systems.
The sample-and-select baseline draws three zero-shot formalizations per
instance and selects one with a reference-free, formalism-specific
selection criterion.  Table~\ref{tab:sampling-summary} reports the selector alongside an oracle which retains any correct sample, i.e., the upper bound for any selection strategy at that budget.  DRAGO beats or ties the
selector in all ten hosted configurations and the \emph{oracle} in nine,
by up to $18.2$ points.  Sampling fails since zero-shot errors are systematic, as testified by
ProverQA-hard/Sonnet where only $7$ of $500$ items exhibit any cross-sample
disagreement.
We then compare against a state-of-the-art system in each of two
formalisms (Table~\ref{tab:dnp}).
Draft-and-Prune~\cite{ni2026draftandpruneimprovingreliabilityautoformalization}
drafts a natural-language \emph{plan} of the
premises' logical structure, converts it into a symbolic program, and
prunes paths that fail to execute.  In its single-path configuration,
DRAGO leads three of the four FOL cells.  The exception, FOLIO/Qwen, is
informative: planning wins where the base model is weakest and the
errors are premise misreadings that a plan fixes before any code
exists, i.e., its advantage tracks zero-shot headroom.  It also reveals a
real gap in DRAGO's coverage: under-constrained formalizations that
execute cleanly and return \emph{undetermined} are invisible to
execution-grounded diagnosis (Limitations and Appendix~\ref{app:failure-analysis}).
CP-SynC~\cite{song2026cpsyncmultiagentzeroshotconstraint} runs
parallel modeling and validation agents and filters candidates through a cascade of targeted repairs. Although it is a strong baseline, its default configuration uses approximately ten times as many LLM calls as DRAGO. DRAGO takes three of the four CSP
configurations, and increasing the trajectories of CP-SynC
does not improve accuracy conditional on solving (Appendix
Table~\ref{tab:cpsync-full}).
Its one win, like the Draft-and-Prune exception, falls on Text2Zinc/Qwen: the
single configuration where DRAGO also loses to plain LLM-diagnostic
repair (Table~\ref{tab:main-results}), possibly because the zero-shot artifacts
are too poor to aim at (Section~\ref{sec:main-results}).

\subsection{Diagnosis Quality}
\label{sec:diagnosis-quality}

\begin{table}[!t]
\centering
\small
\setlength{\tabcolsep}{3.0pt}
\begin{tabular}{@{}lcccccc@{}}
\toprule
& \multicolumn{3}{c}{Qwen3-Coder 30B} & \multicolumn{3}{c}{Gemma 4 31B} \\
\cmidrule(lr){2-4}\cmidrule(lr){5-7}
& Zero & Sys & DRAGO & Zero & Sys & DRAGO \\
\midrule
FOLIO & $60.9$ & $\mathbf{70.3}$ & $64.8$ & $85.0$ & $87.1$ & $\mathbf{87.6}$ \\
ProverQA & $60.8$ & $65.4$ & $\mathbf{68.8}$ & $75.1$ & $75.4$ & $\mathbf{83.7}$ \\
\midrule
CP-Bench & $20.3$ & $27.0$ & $\mathbf{42.9}$ & $43.9$ & $47.6$ & $\mathbf{70.9}$ \\
Text2Zinc & $12.0$ & $\mathbf{20.0}$ & $19.8$ & $20.4$ & $41.8$ & $\mathbf{56.5}$ \\
\bottomrule
\end{tabular}
\caption{Domain-specific systems against DRAGO on the open-source models, in their single-trajectory ($k{=}1$) configurations. Sys: Draft-and-Prune (FOL, top) and CP-SynC (CSP, bottom). \textbf{Bold} marks the best method per benchmark and model.}
\label{tab:dnp}
\end{table}

Accuracy does not show whether the diagnoses themselves are reliable, so we measure them directly on the hosted FOL runs ($5{,}020$ artifacts) and inspect two CSP cases by hand (Appendix~\ref{app:failure-analysis}).  DRAGO raises a blocking finding on $95.3\%$ of the initially wrong artifacts and on $42.6\%$ of the initially correct ones; LLM self-review flags $100\%$ and $90.6\%$.  Deterministic diagnosis is thus a high-recall trigger, and the gate supplies the precision: on initially correct artifacts it rejects $94.6\%$ of the proposed rewrites that would have broken the answer.

\subsection{Pitfalls in Evaluating Formalization Tasks}
\label{sec:measurement-validity}

Across the three formalisms, the reference side of the evaluation required intervention: for FOLIO we adopt a third-party expert-curated relabel~\cite{brunello2026fixingfoliomallsverified}; for Text2Zinc we corrected reference-answer metadata against upstream sources.
On Planetarium, we found two previously undocumented rendering defects that leave $58$/$100$ sampled floor-tile instances, and exactly half of all $6{,}172$ rows of the full split, input--reference inconsistent (Figure~\ref{fig:floor-tile-defect-case}). Repairing the two defects in the problem text recovers $90\%$ (Haiku) and $84\%$ (Sonnet) of the misses on that family. The pattern is not incidental: benchmark references are themselves formalization artifacts, and they exhibit the same silent failure mode investigated in this paper, which we diagnose using our diagnostic methodology. Appendix~\ref{app:benchmark-details} details the defects, the audit, and the corrected release.

%% file: sections/conclusion.tex
\section{Conclusions}
\label{sec:conclusion}

We introduce DRAGO, a repair framework that combines deterministic, reference-free diagnosis with evidence-gated acceptance to improve the translation of natural-language problems into formal specifications. Across three formalisms and four models, it cuts regressions by nearly an order of magnitude ($1.3\%$ vs.\ $11.8\%$) while improving accuracy over both zero-shot generation and LLM self-review. DRAGO also outperforms state-of-the-art per-domain baselines in most configurations with a comparable or lower inference budget. Finally, we uncover two defects in Planetarium and release the corrected split.

\section*{Limitations}

For FOL and CSP, we measured answer accuracy rather than full semantic equivalence: an artifact may return the correct answer while encoding unintended semantics that a different query would expose. The main blind spot, common to all three formalisms, is \emph{execution-blind under-constraint}: an artifact that omits part of the intended problem (a missing axiom, constraint, or goal condition) can still execute cleanly and offers no observable failure to diagnose. These results demonstrate a transferable repair pattern rather than a universal diagnostic system: a new formalism reuses DRAGO's typed-finding interface and acceptance discipline but requires its own diagnostic rules.

%% file: sections/appendix.tex
\section{Appendix}
\label{app:roadmap}

\subsection{Full Method Details}
\label{app:full-method-details}

Let $x$ be a natural-language instance, $p$ a generated formal artifact, and $S(p)$ the executor trace exposed by the target solver stack.  During generation and repair, DRAGO may use $x$, the current artifact $p$, the executor trace $e=S(p)$, and deterministic diagnoses computed from these observable objects.  It does not use a reference formalization, a gold answer, or the final benchmark correctness functional $M(x,p)$ except after the candidate has been returned for scoring.

Each implementation follows the same high-level procedure:
\begin{enumerate}
    \item generate an initial artifact with the zero-shot output contract.
    \item execute the artifact and derive execution-related status information.
    \item compute deterministic reference-free diagnoses $D(x,p,e)$.
    \item ask an LLM to rewrite the artifact using the diagnoses as targeted
    feedback.
    \item execute the candidate rewrite and accept it only if the reference-free
    acceptance checks pass.
    \item stop when no blocking diagnosis remains or when the repair budget is
    exhausted.
\end{enumerate}

The per-formalism diagnosis sources and acceptance rules of this procedure are collected in Table~\ref{tab:diagnostic-instantiations}; default budgets and fallback behavior are listed in Appendix~\ref{app:implementation-details}.

\subsection{Diagnostic and Acceptance Instantiations}
\label{app:diagnostic-rules}

The instantiations below are defined once per formalism.  They are not
selected or modified for a particular benchmark or model: FOLIO and
ProverQA-hard share the complete FOL configuration, Text2Zinc and CP-Bench
share the complete MiniZinc configuration, and the Planetarium subdomains
share the complete PDDL configuration.  The same formalism-level
configurations are used for all four evaluated models.

\paragraph{Rule development protocol.}
The rules were derived empirically, under the same protocol in all three
formalisms, from development data disjoint from every evaluated instance:
we sampled a set of development problems, generated zero-shot
formalizations, executed them, and manually inspected the failures;
recurring failure modes were then distilled into the contract checks,
solver probes, and specification-alignment checks,
together with their acceptance conditions.  For FOL, development problems
were sampled from the FOLIO training split; neither the FOLIO validation
split nor ProverQA-hard contributed any instance, so ProverQA-hard serves
as a fully held-out transfer benchmark for the FOL configuration.  For
PDDL, development problems were sampled from all three domains:
blocksworld and gripper problems from the Planetarium training split, and
floor-tile problems (a domain for which Planetarium provides no training
data and that appears only in the test split) from test-split rows
disjoint from the evaluated sample; the evaluated 300-problem sample
(Appendix~\ref{app:benchmark-details}) shares no instance with the
development set.  For CSP, development problems were drawn from NLP4LP,
hakank, and CSPLib, restricted to
instances that appear in neither Text2Zinc nor CP-Bench; CP-Bench
contributed no development instance.  Once distilled, each configuration
was frozen and applied unchanged to all benchmarks and evaluated models of
its formalism.
Table~\ref{tab:diagnostic-instantiations} summarizes how each formalism
instantiates the three diagnostic classes of the methodology (contract
checks, solver probes, and specification alignment) and the acceptance
conditions. Findings are classified
as \emph{blocking} or \emph{advisory}: advisory findings carry zero weight
and only accompany the repair prompt when a blocking finding has fired, so
the diagnostic potential $w$ of the main text coincides with the total
blocking-finding weight.

\paragraph{Diagnostic potential.}
The diagnostic potential $w$ of Section~\ref{sec:drago} is instantiated
per formalism as a lexicographic potential over (execution status,
blocking findings): in FOL, $w$ counts blocking findings (parse failure,
inconsistent premises, contract violations, underdetermined entailment);
in MiniZinc, $w$ orders non-runnable $\succ$ unsatisfiable $\succ$ number
of semantic-lint findings; in PDDL, $w$ is the total blocking-finding
weight of Table~\ref{tab:diagnostic-instantiations}.  Non-runnable
artifacts are first returned to the runnable region using the ordinary
solver-error channel of Section~\ref{sec:repair-feedback}.  Acceptance is
evaluated on the stage's final candidate, and the previous state is
restored if no candidate strictly reduces $w$.

\begin{table*}[t]
\centering
\setlength{\tabcolsep}{3.0pt}
\begin{tabular}{@{}p{0.11\textwidth}p{0.21\textwidth}p{0.22\textwidth}p{0.21\textwidth}p{0.17\textwidth}@{}}
\toprule
Formalism & Contract checks & Solver probes & Specification alignment & Acceptance \\
\midrule
FOL / SMT-LIB2 &
Single entity sort; premise assertions plus one named conclusion
definition; no solver-control commands; signature vocabulary and arities &
Z3 premise-consistency and entailment-status probes; witness models for
both readings on underdetermination &
Disjunction strength vs.\ exclusivity marking in the problem statement
(advisory) &
Changed, parseable, signature-clean, premise-consistent; corroboration for
undetermined$\rightarrow$\allowbreak definite flips \\
\addlinespace
CSP / MiniZinc &
Data-schema instantiation; required output conventions; static model check &
Deletion probes on UNSAT localize conflicting requirement blocks with the
data they reference &
Semantic lint: quantity granularity and data/unit magnitudes vs.\ the
problem statement &
Passes model check; still runs and solves; no regression to error or UNSAT \\
\addlinespace
PDDL &
Parses and grounds against the official domain file; correct
\texttt{:domain}; declared predicates, arities, and objects; flat
ground-conjunction goal &
Delete-relaxation reachability of goal atoms; bounded complete search
(exhaustion proves unsolvability; timeout yields no finding) &
Count phrases in the statement vs.\ per-category object counts (blocking);
unmatched identifier mentions (advisory) &
Contract holds; strict blocking-weight descent with no new blocking axis; no
goal or object-count weakening \\
\bottomrule
\end{tabular}
\caption{Instantiation of the three diagnostic classes and the acceptance
conditions for each formalism.  Full rule inventories, thresholds, and
defaults are given in the paragraphs below and in the released code.}
\label{tab:diagnostic-instantiations}
\end{table*}

\paragraph{FOL / SMT-LIB2.}
Contract checks verify the fixed output contract of the generated
SMT-LIB2~\cite{BST10}: a single uninterpreted entity sort, premise
assertions plus one named conclusion definition, no solver-control
commands, and predicates drawn from the provided signature with the
declared arities.  Solver probes run Z3~\cite{10.1007/978-3-540-78800-3_24}
to test premise consistency and to compute the entailment status of the
conclusion; the resulting statuses map to labels as \texttt{entailed}
$\rightarrow$ True, \texttt{contradicted} $\rightarrow$ False, and
\texttt{undetermined} $\rightarrow$ Uncertain, while parse errors,
inconsistent premises, and Z3 unknown statuses are tracked separately.
When both the conclusion and its negation are satisfiable, the finding
reports underdetermination together with witness models for each reading.
Alignment checks compare the strength of disjunctions in the script
against exclusivity marking in the problem statement: when the statement
licenses more exclusive disjunctions than the script contains, the finding
names the sentences involved, and in the sentence-aligned case licenses a
deterministic \texttt{or} to \texttt{xor} upgrade that is applied without
an LLM call and remains subject to acceptance.  Alignment findings based
on surface language alone are advisory and never trigger repair by
themselves.

A repair is accepted only if it changes the script, parses, preserves the
signature vocabulary, and keeps the premises consistent.
Undetermined-to-definite flips carry a higher evidential burden: they are
retained only when an independently generated neutral formalization
corroborates the same definite label, up to a narrow false-salvage
exception for contradiction flips.  Concretely, a premise sentence is
treated as an exclusive disjunction when it contains an explicit
exclusivity marker (``but not both'') or a bare
``either \ldots or'' without the inclusive marker ``or both''; a deficit
finding fires when such sentences outnumber the exclusive structures in
the script.  The deterministic upgrade is attempted only when premise
assertions align positionally one-per-sentence (tolerating one trailing
auxiliary assertion), and an LLM repair whose only command-level change is
the same \texttt{or}$\rightarrow$\texttt{xor} upgrade on at most the
cue-licensed assertions is accepted without further corroboration.  For
all other undetermined-to-definite flips, corroboration regenerates an
independent formalization from scratch and accepts the flip only when
that sample is structurally valid and reproduces the same definite label;
the false-salvage exception additionally accepts a False flip when the
neutral sample is valid but remains Uncertain, recovering missed
contradictions without reopening the noisier True flips.

\paragraph{MiniZinc.}
Contract checks verify that the generated model~\cite{1771668.1771709}
instantiates the provided data schema, declares the required output
conventions, and passes MiniZinc's static model check; ordinary
parse/runtime failures are first repaired from solver feedback.  Solver
probes fire on unsatisfiable instances: deterministic block-deletion
probes, in the spirit of deletion-based MUS
extraction~\cite{10.1007/s10817-007-9084-z}, localize the conflicting
requirement blocks and report the parameter values they reference.
Alignment checks run static semantic lint on solved models, comparing
declared structure against the problem statement: quantities described as
countable but declared continuous, and magnitude or unit
mismatches between the instance data and model coefficients.

A candidate repair is accepted only if it passes
\texttt{--model-check-only} and still runs and solves; repairs that
regress to an error or to UNSAT are rejected and the previous runnable
model is restored.

\paragraph{PDDL.}
Contract checks verify the generated problem file~\cite{mcdermott_pddl_1998}
against the official domain file only: the file must parse and ground, name
the required \texttt{:domain}, use declared predicates with the correct
arities and declared objects, and keep the goal in the flat
ground-conjunction fragment; declared objects with no initial fact are
flagged as advisory.  Solver probes are planner-backed: a delete-relaxation
reachability fixpoint over the grounded task proves individual goal atoms
unachievable, with one step of lifted causal backchaining reported as
evidence, and a bounded complete search run in a subprocess proves joint
unsolvability when it exhausts the search space, localized by goal-atom
deletion probes; a search timeout yields no evidence and produces no
finding.  Alignment checks are benchmark-agnostic: count phrases in the
problem statement, matched with a closed lexicon of English numerals, are
compared against the per-category object counts the artifact declares
(blocking), and identifier-like tokens in the statement that never appear
among the declared objects are reported as advisory.  The remaining
silent failures are solvable-but-different artifacts, invisible to
planner-level evidence by construction.

A candidate repair is accepted only under diagnostic descent: it must
satisfy the artifact contract, strictly reduce the total blocking-finding
weight without introducing a blocking finding on a previously clean axis,
and pass conservatism checks against repair-by-weakening---the goal-atom
count may not decrease, and per-category object counts may only move toward
a count stated in the problem statement.  Repair is triggered only by
blocking findings; advisory findings ride along in the repair prompt when a
blocking finding has already fired.  The reference PDDL and the equivalence
outcome are consulted exclusively at final scoring.

\subsection{Prompt Templates}
\label{app:prompt-templates}

The paper uses compact prompt summaries rather than verbatim prompts. Full templates, including all formatting instructions,
are provided in the released codebase.

\begin{table*}[t]
\centering
\setlength{\tabcolsep}{3.0pt}
\begin{tabular}{@{}p{0.12\textwidth}p{0.19\textwidth}p{0.29\textwidth}p{0.29\textwidth}@{}}
\toprule
Domain & Generation output & Generation inputs and constraints & DRAGO repair inputs and constraints \\
\midrule
FOL & SMT-LIB2 script &
Natural-language premises, conclusion, and vocabulary signature.  The prompt
requires one uninterpreted entity sort, premise assertions only, and a named
\texttt{conclusion} Boolean definition. &
Original problem, vocabulary, current SMT-LIB2 script, solver status, and
deterministic FOL diagnoses.  The repair must preserve the output contract and
make only edits licensed by the English problem and diagnostic evidence. \\
\addlinespace
CSP & MiniZinc model &
Problem description, parameter schema, and required output variables.  The
prompt requires a valid MiniZinc model using external \texttt{.dzn} data and
benchmark-specific output conventions. &
Original problem, runnable structured MiniZinc model, data file, and failed
semantic checks.  The repair must keep parameter names, section markers, and
output conventions while fixing the diagnosed semantic failure. \\
\addlinespace
PDDL & PDDL problem file &
Natural-language planning problem.  The prompt requires exactly one PDDL problem
file, not a domain file, plan, or explanation. &
Original planning problem, generated PDDL, and deterministic reference-free
diagnoses.  The repair must preserve the planning domain and return exactly one
complete PDDL problem file. \\
\bottomrule
\end{tabular}
\caption{Summary of the generation and DRAGO repair prompt schemas.  Exact
prompts are released with the codebase.}
\label{tab:prompt-summary}
\end{table*}

The LLM-diagnostic baseline uses the same artifacts but replaces deterministic diagnosis with LLM self-review.  Once an artifact is runnable, one call asks the model to list concrete modeling issues or report that no issue is found; a second call rewrites the artifact using that feedback. The baseline prompt is constrained to the problem statement, current artifact, executor output, and self-review feedback, with no reference formalization or known answer.

\subsection{Benchmark Details}
\label{app:benchmark-details}

\paragraph{FOLIO.}
FOLIO is a first-order-logic natural-language reasoning benchmark~\cite{han2024folionaturallanguagereasoning}. We do not use the original validation labels directly. Subsequent audits found annotation problems in the original benchmark, including incorrect FOL formalizations, ambiguous natural-language sentences, and NLI-label inconsistencies~\cite{brunello2026fixingfoliomallsverified}; a separate agent-evaluation study likewise evaluates on a solver-verified and repaired FOLIO split rather than the raw labels~\cite{ni2026agentifiedassessmentlogicalreasoning}. We therefore use the expert curated validation release of~\cite{brunello2026fixingfoliomallsverified}, where each validation instance was inspected by reviewers with mathematical and domain-specific expertise, cross-checked, and released with corrected formalizations/labels and explanations of the changes.

The curated FOLIO validation split used in our experiments contains \textit{73 story rows} and \textit{202 flattened conclusion examples}. The label distribution is True 73, False 54, and Uncertain 75. Each example contains natural-language premises, one conclusion, a curated vocabulary/signature, and a gold label in $\{\mathrm{True},\mathrm{False},\mathrm{Uncertain}\}$. Generated SMT-LIB2~\cite{BST10} is executed by Z3~\cite{10.1007/978-3-540-78800-3_24} and mapped back to this label space using the entailment / contradiction / underdetermination checks described above.

\paragraph{ProverQA-hard.}
ProverQA is a synthetic first-order-logic reasoning benchmark generated by ProverGen, a pipeline that combines LLM-generated stories with symbolic-prover validation to produce natural-language premises, formal logical structure, and reasoning traces~\cite{qi2025large}. We use the hard split because it is the least saturated setting in ProverQA and is intended to stress multi-step FOL reasoning rather than surface-level entailment. It therefore complements FOLIO: FOLIO provides expert-curated human-written logical reasoning examples, while ProverQA-hard provides a larger controlled set with symbolic construction and higher reasoning depth. The ProverQA-hard split used in our experiments contains \textit{500 examples} with labels distributed as following: True 143, False 183, and Uncertain 174. We normalize each item into the same schema used for FOLIO: natural-language premises, one conclusion, a relation/constant signature extracted offline from the dataset's reference FOL annotations (the annotations themselves are never shown), and a held-out reference label used only for scoring purposes.

\paragraph{Text2Zinc.}
Text2Zinc is a cross-domain benchmark for translating natural-language optimization and satisfaction problems into MiniZinc~\cite{singirikonda2025text2zinccrossdomaindatasetmodeling}. We chose it because it combines optimization and satisfaction cases in a common solver-agnostic modeling language and because its instances come from multiple problem sources rather than from a single template family. The verified Text2Zinc-style file contains 110 instances: 55 minimization, 45 maximization, and 10 satisfaction problems. The source metadata is \texttt{nlp4lp} (65 instances), \texttt{hakank} (22 instances)~\cite{hakankgithub}, \texttt{csplib} (11 instances)~\cite{csplib}, \texttt{complexor} (7 instances), and \texttt{lpwp} (5 instances); NLP4LP is the optimization dataset introduced with OptiMUS~\cite{ahmaditeshnizi2024optimus}. Instances include natural-language description, MiniZinc model reference executables (not always present), DZN data, parameters, expected output variables, and a reference output dictionary. Before evaluation we performed a lightweight relabeling pass on the reference answers. Unlike the FOLIO curation, this was not a new expert semantic audit of each natural-language problem. The observed issues were straightforward answer metadata errors, most often a final objective value that did not match the released model/data or the original source artifact. We corrected these by checking the upstream source such as NLP4LP. Optimization cases are scored by comparing the generated MiniZinc solution objective against the corrected reference objective, while satisfaction cases are scored by solver feasibility and the declared output contract.

\paragraph{CP-Bench.}
CP-Bench is a constraint-programming benchmark for natural-language-to-model evaluation~\cite{michailidis2026dcpbenchopenevaluatingllmsconstraint}. The original benchmark offers a larger collection of questions, but we restrict our experiments to the verified subset to reduce label noise and avoid scoring against uncertain references. The verified CP-Bench subset used in our experiments contains 63 instances: 45 satisfaction, 13 minimization, and 5 maximization problems. We adapt each instance into the same MiniZinc-facing problem schema used for Text2Zinc, run generated MiniZinc models with the same solver interface, and score satisfaction cases by feasibility and optimization cases by the corrected objective/output contract.

\paragraph{Planetarium.}
Planetarium is a text-to-PDDL benchmark for evaluating semantic correctness in generated planning problem files~\cite{zuo-etal-2025-planetarium}. The original test split contains 15,943 instances over the three supported domains: 4,993 \texttt{blocksworld}, 6,172 \texttt{floor-tile}, and 4,778 \texttt{gripper}. Because running the full test split across all models, baselines, and repair conditions exceeds our available compute budget, we evaluate on a fixed seeded sample of 300 problems, 100 per domain, drawn once and shared by every model and method, and report both per-subdomain and overall accuracy. The sample is reproducible byte-for-byte: within each domain, rows are sorted by problem name, which is stable under dataset shuffling, and 100 problems are drawn without replacement using Python's \texttt{random.Random(42)}, with one shared generator across the three domains in the order blocksworld, floor-tile, gripper.

\begin{figure}[t]
\centering
\begin{tabular}{@{}p{0.97\columnwidth}@{}}
\toprule
\textbf{Question as shipped} (floor-tile, $1{\times}3$ grid, two robots) \\
\midrule
\ttfamily\frenchspacing
You have 2 robots.\newline
You have 3 tiles.\newline
You have 3 colors.\newline
\textbf{Tile tile1 is is to the right of tile tile2.}\newline
\textbf{Tile tile2 is is to the right of tile tile3.}\newline
The robot robot1 has color color1.\newline
\textbf{The robot is at tile robot1.}\newline
The robot robot2 has color color2.\newline
\textbf{The robot is at tile robot2.}\newline
Color color1 is available.\newline
Color color2 is available.\newline
Color color3 is available.\newline
Your goal is to have the following:\newline
Tile tile1 should be painted with color color1.\newline
Tile tile2 should be painted with color color1.\newline
Tile tile3 should be painted with color color1. \\
\midrule
\textbf{Reference PDDL} (init excerpt) \\
\midrule
\ttfamily\frenchspacing
(robot-at robot1 tile1)\quad(robot-at robot2 tile2)\newline
(right tile2 tile1)\qquad\ \,(right tile3 tile2) \\
\bottomrule
\end{tabular}
\caption{The smallest sampled floor-tile instance, verbatim, with the
defective sentences in bold.  \emph{Defect 1 (destroyed robot positions):}
``The robot is at tile robot1.'' fills the tile slot with the robot's own
name; the true positions, \texttt{(robot-at robot1 tile1)} and
\texttt{(robot-at robot2 tile2)}, appear nowhere in the question.
\emph{Defect 2 (reversed adjacency):} the reference states
\texttt{(right tile2 tile1)}---under the shipped domain file's action
semantics, tile2 is to the right of tile1---while the question asserts
the opposite, ``Tile tile1 is\ldots to the right of tile tile2.''  The
duplicated ``is is'' is verbatim.  A model that reads this text faithfully
places no robot correctly and builds the mirrored row.}
\label{fig:floor-tile-defect-case}
\end{figure}

\paragraph{Floor-tile rendering defects.}
Zero-shot floor-tile accuracy was strikingly poor for both hosted models
(Haiku $22/100$, Sonnet $40/100$), far below the same models' blocksworld
and gripper results, so we manually inspected every missed instance.  The misses split sharply
along the benchmark's two template families: on the $58$ sampled problems
whose text states the grid layout sentence-by-sentence, both models score
$0/58$, against $22/42$ (Haiku) and $40/42$ (Sonnet) on the rows that
describe the layout structurally.  Reading the defective rows against
their reference PDDL revealed two errors in the question rendering
(Figure~\ref{fig:floor-tile-defect-case} shows a complete instance
exhibiting both); each was verified byte-for-byte against the upstream
dataset release, independently of our data pipeline.  \emph{First, robot positions are
destroyed.}  The template renders every robot position as ``The robot is
at tile robot1.''. The tile slot is filled with the robot's own name, so
the true position, present in the reference PDDL, appears nowhere in the
question.  \emph{Second, every adjacency sentence states the reverse
relation.}  The shipped domain file fixes the predicate convention through
its actions (\texttt{paint-up} paints \texttt{?y} where \texttt{(up ?y
?x)}, so the first argument is the tile above); under that reading, all
$1{,}461$ rendered right/above sentences across the sampled floor-tile
rows contradict their row's reference PDDL, and all $1{,}461$ match the
swapped reading.  A deterministic, LLM-free sweep of the full floor-tile
test split ($6{,}172$ rows; dataset snapshot), detecting
each defect independently per row, confirms corpus-wide prevalence: the
dataset renders every problem at both initial-state abstraction levels,
and the defective family coincides exactly with its
\texttt{init\_is\_abstract=false} rows---all $3{,}086$ of them carry both
defects
($12{,}311/12{,}311$ robot-position sentences broken;
$68{,}716/68{,}716$ relation sentences consistent only with the swapped
reading), no row carries exactly one defect or an unparsable rendering,
and the $58/100$ sample split is verified row-by-row against the full
split by content hash.  The models' wrong outputs are faithful readings of this
defective text: they transcribe the broken position sentence verbatim and
build exactly the mirrored grid the sentences describe.  The defects are not
fixable by prompting: stating the reversed convention in the prompt raises
Haiku's defective family from $0$ to $47$ correct but collapses the clean
family from $22$ to $5$, and correct worked examples collapse it from $22$
to $12$.

\paragraph{Why the defects exist and why they went unnoticed.}
Both errors are implementation-level template bugs, not flaws in the
benchmark's design.  Planetarium's questions are generated automatically,
by filling and procedurally combining handcrafted sentence
templates from the reference PDDL~\cite{zuo-etal-2025-planetarium}.  The
first defect is a slot-filling error---the position slot receives the
robot's identifier instead of its tile---and the second swaps the
argument-order convention of the adjacency predicates relative to the
domain file; a duplicated word (``Tile tile1 is is to the right
of\,\ldots'') appears in every one of the split's $68{,}716$ relation
sentences, locating both the duplication and, plausibly, the swap in the
template itself.  The defects could persist because nothing in the
benchmark pipeline reads the text: rendering and scoring are both
automatic, and floor-tile appears only in the test split, so no training
run ever surfaced these renderings.
This is the failure mode this paper studies, occurring on the reference
side: the defective questions render and score without any error signal
while encoding the wrong semantics.
The audit is an offline activity with reference
access; the repair method remains reference-free at test time.  Neither defect is documented in the
benchmark paper (which attributes floor-tile difficulty to the domain
being less common), in its public issue tracker, or in follow-up
evaluations.  All
four evaluated models and every method score $0/58$ on these rows in every
run, so they act as a fixed offset: the achievable score is $42/100$ on
floor-tile and $242/300$ overall.

\paragraph{Repaired-rendering ablation and reporting policy.}
Repairing the $58$ defective renderings (restoring robot positions from
each row's reference problem---the only step that needs the reference---and
un-mirroring the adjacency sentences, a purely textual edit) and rerunning
the campaign raises zero-shot Planetarium accuracy by $+17.3$ points for
Haiku and $+16.3$ for Sonnet, with the ranking of arms and DRAGO's
low-regression safety profile unchanged: the ablation quantifies the
defects' cost without altering any comparative conclusion.  On the $58$
previously failing rows themselves, zero-shot generation from the repaired
text recovers $52/58$ (Haiku) and $49/58$ (Sonnet) under identical
prompts, extraction, and evaluator.  A full input-side $2{\times}2$
factorial confirms the attribution per defect: repairing only the robot
positions recovers $4/58$ for both models (a correct position inside a still-mirrored grid rarely
survives scoring), and repairing only the adjacency sentences recovers
$6/58$ (Haiku) and $5/58$ (Sonnet), so each defect is independently
sufficient to make a row score zero and both must be repaired jointly.
The same two repair operators, applied to the full test split, define a
corrected release: all $3{,}086$ defective rows are patched (robot
positions restored from each row's \texttt{(:init)} section, so a goal
atom can never override the initial position; relation sentences
un-swapped and the duplicated ``is is'' normalized), the $3{,}086$ clean
rows remain byte-identical, and the build fails unless re-running the
independent census audit on the patched split reports zero defective
sentences, which it does: $12{,}311$ position sentences restored and all
$68{,}716$ relation sentences now domain-consistent with their row's
reference.  The patched split is released with the codebase.
Following
the principle of Section~\ref{sec:measurement-validity}, the shipped
benchmark remains the primary condition: unlike the FOLIO and Text2Zinc
corrections, which defer to external ground truth, this rewrite of $58\%$
of a subdomain's task inputs was authored by us from the reference PDDL,
so it is reported only as an ablation and the primary numbers stay
comparable with prior and future work.

\subsection{Implementation Details}
\label{app:implementation-details}

\paragraph{Models and API settings.}
The paper runs use \texttt{claude-haiku-4-5-20251001}~\cite{anthropic2025claudehaiku45systemcard} and \texttt{claude-sonnet-4-6}~\cite{anthropic2026claudesonnet46systemcard}.  The local project declares \texttt{anthropic>=0.107.1}; every Anthropic request sets the model, system prompt, user message, and \texttt{max\_tokens=8192}.  The request payload does not set \texttt{temperature}, \texttt{top\_p}, or \texttt{top\_k}, so hosted-model runs use provider-default sampling for the pinned model identifiers above; run-to-run variability is instead reported empirically through repeated runs.  The open-source track uses \texttt{Qwen/Qwen3-Coder-30B-A3B-Instruct}~\cite{yang2025qwen3technicalreport} and \texttt{google/gemma-4-31B-it}~\cite{gemmateam2026gemma4technicalreport} served using the vLLM engine, on a single node of NVIDIA 4xA100. Qwen3-Coder uses model-native sampling with \texttt{temperature=0.7}, \texttt{top\_p=0.8}, \texttt{top\_k=20}, and per-run sampling seeds (the vLLM request seed; seeds 1--5) for the multi-seed significance sweep.  Gemma likewise uses model-native sampling (\texttt{temperature=1.0}, \texttt{top\_p=0.95}, \texttt{top\_k=64}) with per-run sampling seeds (1--5; 1--3 for ProverQA-hard).  The open-model PDDL campaign uses three sampling seeds (1--3) per model; within each seed, all three arms run against one vLLM server and the repair arms consume that seed's own zero-shot artifacts, so the PDDL comparisons are paired by construction.  The Planetarium subdomain breakdown (Appendix Table~\ref{tab:pddl-subdomains}) is computed over the same seeds as the overall score.

\paragraph{FOL.}
The final FOL harness repeats FOLIO five times and ProverQA-hard three times. FOL DRAGO defaults to one semantic repair round, corroboration enabled, a two-sample cap for ordinary structurally rejected repair samples, and a four-sample cap for parse-error or inconsistent-premise cases.  Diagnosis and scoring run Z3 5.0.0 (\texttt{z3-solver} 5.0.0.0) on the host CPU; hosted-model generation and repair execute provider-side.

\paragraph{MiniZinc.}
The zero-shot MiniZinc baseline runs generation followed by solve/evaluation. The solver mode is \texttt{auto}, which selects CP/integer or LP-friendly solver candidates according to the model.  The current DRAGO defaults are two ordinary runnable-repair attempts, two UNSAT diagnosis-repair rounds, a 10-second UNSAT probe timeout and one semantic repair round.  All MiniZinc runs use MiniZinc 2.9.7 with the bundled Gecode 6.3.0 and HiGHS 1.14.0 solvers.  The hosted-model CSP campaign repeats each model--benchmark configuration over five independent paired runs (the primary sweep plus four replications executed afterwards under identical settings), each including the one-round LLM-diagnostic replay of Appendix~\ref{app:statistical-analysis}.

\paragraph{PDDL.}
The zero-shot baseline runs schema-aware generation followed by Planetarium
evaluation; all three arms share the same generation prompt and the same
plain response extraction.  The LLM-diagnostic baseline uses the same self-review budget as
the other formalisms.  PDDL DRAGO uses the planner-backed diagnosis stack of
Appendix~\ref{app:diagnostic-rules} with a one-round repair budget, up to three repair samples per round,
and a 10-second bounded-search probe timeout; diagnosis itself is LLM-free.  The hosted-model PDDL campaign comprises three independent paired runs per model.  The planner stack uses pyperplan for parsing, grounding, and
bounded search, with domain files taken from the installed Planetarium
package; problem-level \texttt{(:requirements ...)} declarations are
stripped before parsing, which is legal PDDL that the pyperplan parser does
not support and does not change task semantics.

\subsection{Motivating Silent-Failure Taxonomy}
\label{app:motivation-silent-failures}

Table~\ref{tab:silent-failure-taxonomy} separates zero-shot failures into ordinary execution failures and silent semantic failures.  A row is counted as \emph{non-runnable wrong} when the generated artifact fails to parse, execute, or reach a meaningful solver/evaluator outcome.  A row is counted as \emph{runnable wrong} when it parses and executes far enough to produce a solver, prover, or planner/equivalence outcome, but the benchmark answer is incorrect.  FOL includes repeated stochastic runs, CSP uses the retained primary-run sweeps, and PDDL uses the two hosted primary paired runs on the seeded 300-problem Planetarium sample.

\begin{table*}[t]
\centering
\setlength{\tabcolsep}{4.0pt}
\resizebox{\textwidth}{!}{%
\begin{tabular}{@{}llrrrrr@{}}
\toprule
Formalism & Retained benchmarks & $N$ & Correct & Incorrect
& Non-runnable wrong & Runnable wrong / wrong \\
\midrule
FOL & FOLIO, ProverQA-hard & 5020 & 3932 & 1088 & 131 & $957/1088$ ($88.0\%$) \\
CSP & Text2Zinc, CP-Bench & 346 & 126 & 220 & 134 & $86/220$ ($39.1\%$) \\
PDDL & Planetarium & 600 & 413 & 187 & 0 & $187/187$ ($100.0\%$) \\
\midrule
All & Retained outputs & 5966 & 4471 & 1495 & 265 & $1230/1495$ ($82.3\%$) \\
\bottomrule
\end{tabular}%
}
\caption{Zero-shot failure taxonomy over the retained saved outputs.  The
rightmost column reports failures that are runnable but semantically wrong,
which ordinary solver-error feedback would not expose as syntax or runtime
errors.}
\label{tab:silent-failure-taxonomy}
\end{table*}

This taxonomy motivates the diagnostic layer directly.  In FOL, most zero-shot failures are not parser or runtime failures but executable formalizations that produce the wrong entailment label; in PDDL, \emph{every} hosted-model zero-shot failure parses and evaluates cleanly while failing semantic equivalence; this remains true after excluding the $116$ defective floor-tile rows ($58$ per model, Appendix~\ref{app:benchmark-details}), whose failures are runnable-wrong by construction.  CSP exposes more ordinary MiniZinc execution failures, but even there $39.1\%$ of zero-shot failures execute to a solver outcome while encoding the wrong model. Execution success is therefore not a reliable proxy for formalization correctness.

\subsection{Interpreting the Paired McNemar Test}
\label{app:mcnemar-example}

McNemar's test compares two methods on the same instances.  For example,
suppose zero-shot and DRAGO are evaluated on $20$ instances with the following
paired outcomes: both methods correct on $7$ instances, DRAGO correct where
zero-shot is wrong on $8$ (fixes), DRAGO wrong where zero-shot is correct on
$1$ (break), and both methods wrong on $4$.

Let $f$ be the number of fixes, $b$ the number of breaks, and $N$ the number
of paired instances.  The paired accuracy difference is
\[
\Delta\mathrm{Acc}=\frac{f-b}{N}.
\]
Here, $\Delta\mathrm{Acc}=(8-1)/20=0.35$.
McNemar's exact two-sided test uses only the $f+b=9$ discordant pairs.  Under
the null hypothesis that either method is equally likely to be correct on a
discordant instance, the number of fixes follows a $\mathrm{Binomial}(9,0.5)$
distribution, giving
\[
p=2\sum_{k=0}^{1}\binom{9}{k}(0.5)^9=0.039.
\]
Thus the test asks whether the observed gain comes from fixing more instances
than it breaks; it does not use the $11$ instances on which the methods agree.

\subsection{Statistical Analysis}
\label{app:statistical-analysis}

For each benchmark/model/method configuration we report mean accuracy and sample standard deviation across independent runs.  Repeated stochastic runs stabilize estimates, but they are not treated as additional benchmark items.  All repair methods are evaluated at the one-round budget. The hosted CSP sweep comprises five independent paired runs per model and benchmark; the hosted CSP cells of Table~\ref{tab:main-results} average all five runs, with the LLM-diagnostic baseline entering through its one-round replay in each run.  The direct DRAGO-versus-baseline CSP discordances in Table~\ref{tab:domain-mcnemar} are drawn from one complete paired run with both repair arms at the one-round budget; the remaining hosted cells of Table~\ref{tab:domain-mcnemar} pool discordant pairs over the repeated paired runs (five for CSP, three for PDDL, and the three common runs for FOL).  Because pooled counts reuse the same benchmark instances across runs, the per-run analyses below accompany them, and every DRAGO comparison reported as significant in Table~\ref{tab:domain-mcnemar} is also significant in each individual run.  Table~\ref{tab:main-results} reports mean and standard-deviation accuracy only; the benchmark-level McNemar outcomes~\cite{McNemar1947} are summarized here and in Table~\ref{tab:opensource-mcnemar}.  For Qwen3-Coder, DRAGO is significant in every available seed for CP-Bench ($5/5$) and ProverQA-hard ($3/3$); FOLIO and Text2Zinc are pooled-significant but per-seed fragile ($1/5$ significant seeds each).  On PDDL (three engine seeds per open model), Qwen's DRAGO gain is marginal: nominally $p=.041$ when pooling across seeds, but not significant in any individual seed, so we describe it as safe and slightly positive rather than as a win. Against LLM-diagnostic repair, Qwen3-Coder DRAGO is robust on ProverQA-hard ($3/3$) and on PDDL (pooled $p=2.0\times10^{-12}$), borderline on CP-Bench ($3/5$), not significant on FOLIO, and worse on Text2Zinc.  For Gemma, DRAGO against zero-shot is significant in every available seed for PDDL ($3/3$), Text2Zinc ($5/5$), ProverQA-hard ($3/3$), and CP-Bench ($5/5$); FOLIO is significant in only $2/5$ seeds, reflecting its narrow $2.6$-point headroom.  Against LLM-diagnostic repair, Gemma's DRAGO is robust on PDDL (pooled $p=4.9\times10^{-8}$), Text2Zinc ($5/5$), and ProverQA-hard ($3/3$), not significant on FOLIO ($0/5$), and significant in only $1/5$ CP-Bench seeds despite a low pooled $p$.  The open-model PDDL means also contain the one subdomain where the baseline leads DRAGO, Gemma blocksworld ($98.7$ vs.\ $94.7$); the baseline buys that lead with heavy damage on gripper ($73.3\rightarrow55.3$).  For paired significance tests, each item contributes a binary correctness value for each method.  We use exact two-sided McNemar tests on the discordant pairs: \emph{fixes} are items where DRAGO is correct and the comparator is wrong, and \emph{breaks} are items where DRAGO is wrong and the comparator is correct. Concordant items do not affect the McNemar statistic.  Table~\ref{tab:paired-effect-sizes} complements the primary tests with paired accuracy gains and confidence intervals; its CSP and PDDL rows use the primary run, and the domain-level point estimates of all hosted runs fall inside the reported intervals (CSP: $+24.9$ to $+27.7$ points for Haiku, $+15.0$ to $+19.1$ for Sonnet; PDDL: $+2.3$ to $+3.3$ and $+0.3$ to $+1.0$).

The main paper reports domain-level tests because several individual benchmark splits are relatively small.  This merges distinct benchmark items within a domain, such as Text2Zinc with CP-Bench for CSP, while preserving item pairing. This is different from naively pooling repeated runs over the same fixed instances.  For FOL, where the final sweep has multiple stochastic runs, we merge FOLIO and ProverQA-hard within each common run and report whether the domain-level result is significant in every run.  For both FOL models, the DRAGO comparisons in Table~\ref{tab:domain-mcnemar} are significant in $3/3$ common runs.  For the hosted CSP sweep we apply the same discipline over its five paired runs: merging Text2Zinc and CP-Bench within each run, DRAGO versus zero-shot is significant at $p<.0001$ in every run for both models, with per-run discordances ranging from $45/2$ to $50/2$ for Haiku and from $30/4$ to $37/4$ for Sonnet; the LLM-diagnostic baseline's own gain over zero-shot is likewise significant in every run for both models (largest per-run $p{=}.043$).  The hosted PDDL sweep likewise comprises three independent paired runs per model.  For Haiku, DRAGO versus zero-shot is significant in every individual run ($p{=}.039$, $.006$, $.039$; pooled $27/3$); for Sonnet the discordance mass at the measurement ceiling is too small for per-run significance ($1/0$, $3/0$, $2/0$; pooled $6/0$, $p_{\mathrm{Holm}}{=}.063$).  Head-to-head against the baseline, the pooled PDDL discordances are $53/35$ for Haiku ($p{=}.069$; no individual run significant) and $92/2$ for Sonnet ($p{<}.0001$ in every run).

Table~\ref{tab:domain-mcnemar} also reports LLM-diagnostic repair relative to zero-shot.  This is not a head-to-head test against DRAGO, but it shows the repair profile induced by ungrounded self-review from the same starting point. The direct statistical comparison between repair methods is the rightmost block, DRAGO vs. LLM-diagnostic repair.  The two DRAGO comparison blocks are adjusted jointly using Holm--Bonferroni across the $12$ pre-specified hosted-model DRAGO comparisons.

\begin{table*}[t]
\centering
\setlength{\tabcolsep}{2.8pt}
\begin{tabular}{@{}llcccccc@{}}
\toprule
& & \multicolumn{2}{c}{LLM-diag. vs. Zero} &
\multicolumn{2}{c}{DRAGO vs. Zero} &
\multicolumn{2}{c}{DRAGO vs. LLM-diag.} \\
\cmidrule(lr){3-4}\cmidrule(lr){5-6}\cmidrule(lr){7-8}
Domain & Model & Fix/Break & $p$ & Fix/Break & $p_{\mathrm{Holm}}$ & Fix/Break & $p_{\mathrm{Holm}}$ \\
\midrule
CSP & Haiku & $151/20$ & $<.0001$ & $232/10$ & $<.0001$ & $25/8$ & $.018$ \\
CSP & Sonnet & $131/41$ & $<.0001$ & $169/20$ & $<.0001$ & $23/9$ & $.060$ \\
PDDL & Haiku & $42/36$ & $.572$ & $27/3$ & $<.0001$ & $53/35$ & $.069$ \\
PDDL & Sonnet & $6/90$ & $<.0001$ & $6/0$ & $.063$ & $92/2$ & $<.0001$ \\
FOL & Haiku & $148/148$ & $1.00$ & $275/29$ & $<.0001$ & $311/65$ & $<.0001$ \\
FOL & Sonnet & $127/201$ & $<.0001$ & $253/30$ & $<.0001$ & $359/62$ & $<.0001$ \\
\bottomrule
\end{tabular}
\caption{Domain-level exact McNemar tests, with all repair methods at the
one-round budget.  Fix/Break counts pool the discordant pairs over the
repeated hosted runs (five CSP runs, three PDDL runs, three common FOL
runs), except the CSP cells of the rightmost column, which come from the
single complete paired run; per-run tests are reported in the accompanying
text (Appendix~\ref{app:statistical-analysis}).  In each comparison, Fix/Break gives the discordant pairs
where the left method is correct and the right method is wrong, or vice
versa; a significant $p$ therefore also flags harmful effects, as in the
PDDL/Sonnet LLM-diagnostic column, where the baseline breaks $90$ instances
against $6$ fixes.  The two DRAGO comparison columns report
Holm--Bonferroni-adjusted values; the LLM-diagnostic-vs-zero column reports
unadjusted exact values.}
\label{tab:domain-mcnemar}
\end{table*}

\begin{table*}[t]
\centering
\setlength{\tabcolsep}{5.0pt}
\begin{tabular}{@{}llcc@{}}
\toprule
Domain & Model & Paired accuracy gain (pp) & $95\%$ CI \\
\midrule
CSP & Haiku & $+24.9$ & $[17.9,\ 32.4]$ \\
CSP & Sonnet & $+16.2$ & $[10.4,\ 22.0]$ \\
PDDL & Haiku & $+2.3$ & $[0.7,\ 4.3]$ \\
PDDL & Sonnet & $+0.3$ & $[0.0,\ 1.0]$ \\
FOL & Haiku & $+11.7$ & $[9.6,\ 13.8]$ \\
FOL & Sonnet & $+10.6$ & $[8.5,\ 12.7]$ \\
\bottomrule
\end{tabular}
\caption{Paired effect sizes for the primary DRAGO-vs-zero-shot comparison,
in percentage points.  Intervals are $95\%$ percentile paired-bootstrap
intervals with $4{,}000$ replicates over the matched benchmark instances.  For
FOL, we use the three common runs, resample runs and instance units in two
stages, and resample FOLIO at the story level.  CSP and PDDL rows use the primary
paired run; the point estimates of the additional hosted runs
fall inside the reported intervals (see text).}
\label{tab:paired-effect-sizes}
\end{table*}

\subsection{Open-Source Paired Analyses}
\label{app:opensource-mcnemar}

Table~\ref{tab:opensource-mcnemar} gives the corresponding paired analyses for
Qwen3-Coder and Gemma.  Because repeated runs reuse the same benchmark
instances, pooled discordances across seeds would overstate the effective sample
size.  We therefore use the per-seed significance rate, rather than a pooled
$p$-value, as the inferential summary.  Fix/Break counts remain useful for
describing the direction and frequency of repair transitions across the saved
runs.

\begin{table*}[t]
\centering
\setlength{\tabcolsep}{3.1pt}
\resizebox{\textwidth}{!}{%
\begin{tabular}{@{}llcccc@{}}
\toprule
& & \multicolumn{2}{c}{Qwen3-Coder 30B} &
\multicolumn{2}{c}{Gemma 4 31B} \\
\cmidrule(lr){3-4}\cmidrule(lr){5-6}
Domain & Benchmark & DRAGO vs. Zero & DRAGO vs. LLM-diag. &
DRAGO vs. Zero & DRAGO vs. LLM-diag. \\
\midrule
PDDL & Planetarium & $15/5;\ 0/3$ & $97/22^{\dagger}$ &
$80/6;\ 3/3$ & $138/61^{\dagger}$ \\
\midrule
CSP & Text2Zinc & $84/41;\ 1/5$ & $61/74;\ 0/5$ &
$209/10;\ 5/5$ & $124/26;\ 5/5$ \\
    & CP-Bench & $88/17;\ 5/5$ & $74/18;\ 3/5$ &
$58/7;\ 5/5$ & $34/12;\ 1/5$ \\
\midrule
FOL & ProverQA-hard & $128/8;\ 3/3$ & $225/90;\ 3/3$ &
$130/2;\ 3/3$ & $170/28;\ 3/3$ \\
    & FOLIO & $61/21;\ 1/5$ & $135/123;\ 0/5$ &
$34/8;\ 2/5$ & $59/41;\ 0/5$ \\
\bottomrule
\end{tabular}%
}
\caption{Open-source paired McNemar analyses, grouped by formalism.  Each cell
reports \emph{Fix/Break; significant runs/available runs}.  Fix/Break sums the
discordant paired outcomes over saved runs and is descriptive; significance is
assessed separately by an exact two-sided McNemar test in each seed at
$\alpha=0.05$.  With the exception of the Planetarium DRAGO vs.\
LLM-diagnostic comparison ($^{\dagger}$), where we pool across
seeds ($p=2.0\times10^{-12}$ for Qwen3-Coder, $p=4.9\times10^{-8}$ for Gemma),
we do not report pooled $p$-values because repeated runs reuse the same
benchmark instances.}
\label{tab:opensource-mcnemar}
\end{table*}

The transition counts also give a direct comparison between the two repair loops relative to zero-shot.  A repair method can improve accuracy by fixing initial errors, but it can also trade those fixes against regressions on initially correct formalizations.  DRAGO's domain-level profile has the more favorable balance: fixes substantially outnumber breaks against both zero-shot and LLM-diagnostic repair in CSP, FOL, and PDDL.

\subsection{Additional Results}
\label{app:additional-results}

Figure~\ref{fig:fol-transitions} showed the correctness-transition breakdown for Haiku on FOL. The breakdown is extended to Planetarium for both hosted models (Table~\ref{tab:anthropic-transitions-pddl}) and to the open-source track (Table~\ref{tab:opensource-transitions}); for Sonnet on FOL, the same asymmetry is visible in the domain-level fix/break counts of Table~\ref{tab:domain-mcnemar} ($253/30$ for DRAGO versus $127/201$ for LLM-diagnostic repair against zero-shot).

\paragraph{Planetarium subdomain accuracy.}
Table~\ref{tab:pddl-subdomains} reports the per-subdomain accuracy behind
the Planetarium column of Table~\ref{tab:main-results}.  The floor-tile
column is dominated by the $58$ input--reference-inconsistent defective instances
(Appendix~\ref{app:benchmark-details}), so its achievable maximum is $42$.

\begin{table}[t]
\centering
\setlength{\tabcolsep}{4.0pt}
\begin{tabular}{@{}lrrrr@{}}
\toprule
Method & Blocks & Floor & Gripper & Overall \\
\midrule
\multicolumn{5}{@{}l}{\emph{Haiku 4.5}} \\
Zero & $83.0$ & $22.0$ & $69.0$ & $58.0$ \\
LLM-diag. & $80.0$ & $20.0$ & $76.0$ & $58.7$ \\
DRAGO & $90.0$ & $21.0$ & $70.0$ & $60.3$ \\
\midrule
\multicolumn{5}{@{}l}{\emph{Sonnet 4.6}} \\
Zero & $100.0$ & $40.0$ & $99.0$ & $79.7$ \\
LLM-diag. & $97.0$ & $30.0$ & $83.0$ & $70.0$ \\
DRAGO & $100.0$ & $40.0$ & $100.0$ & $80.0$ \\
\midrule
\multicolumn{5}{@{}l}{\emph{Qwen3-Coder 30B}} \\
Zero & $42.0$ & $0.0$ & $3.7$ & $15.2$ \\
LLM-diag. & $14.0$ & $1.7$ & $8.3$ & $8.0$ \\
DRAGO & $45.3$ & $0.3$ & $3.3$ & $16.3$ \\
\midrule
\multicolumn{5}{@{}l}{\emph{Gemma 4 31B}} \\
Zero & $93.3$ & $17.0$ & $73.3$ & $61.2$ \\
LLM-diag. & $98.7$ & $28.7$ & $55.3$ & $60.9$ \\
DRAGO & $94.7$ & $32.0$ & $81.7$ & $69.4$ \\
\bottomrule
\end{tabular}
\caption{Planetarium accuracy by subdomain ($100$ instances each).
Hosted-model rows report one run; open-model
rows are means over three seeded runs.
}
\label{tab:pddl-subdomains}
\end{table}

\paragraph{Hosted-model PDDL correctness transitions.}
\label{app:anthropic-transitions}
Figure~\ref{fig:fol-transitions} reports correctness transitions only for FOL with Haiku.  Table~\ref{tab:anthropic-transitions-pddl} extends the same breakdown to Planetarium for both hosted models, using the primary paired run behind Table~\ref{tab:main-results}.  Each row decomposes the $100$ instances of one subdomain/model/method cell into the four initial$\rightarrow$final buckets R$\rightarrow$R, R$\rightarrow$W, W$\rightarrow$R, and W$\rightarrow$W, where R denotes a correct artifact and W an incorrect or unscored artifact; both repair arms are paired on the same saved zero-shot artifacts, so transitions are measured against a shared initial state.  The CSP benchmarks are excluded because initial correctness is not recoverable from their saved runs: CP-Bench's feasibility-based scoring is computed only post-hoc on the final artifact, and the retained Text2Zinc attempt logs score intermediate artifacts with an attempt-time scorer that does not reproduce the final evaluator.

The hosted models reproduce the asymmetry of Figure~\ref{fig:fol-transitions} on PDDL.  Over the $300$ paired instances, DRAGO breaks a single initially correct instance for Haiku and none for Sonnet, while the LLM-diagnostic baseline breaks up to $31/300$ (Sonnet), turning a $79.7\%$ zero-shot into a net $-29$.  The subdomain split localizes those Sonnet regressions across gripper ($17$ breaks) and floor-tile ($11$ breaks), whereas DRAGO's only movement on Sonnet is one gripper fix.

\begin{table*}[t]
\centering
\setlength{\tabcolsep}{4.0pt}
\begin{tabular}{@{}lllrrrrr@{}}
\toprule
Subdomain & Model & Method &
R$\rightarrow$R & R$\rightarrow$W & W$\rightarrow$R & W$\rightarrow$W & Net \\
\midrule
blocksworld & Haiku 4.5 & DRAGO & 83 & 0 & 7 & 10 & $+7$ \\
            & Haiku 4.5 & LLM-diag. & 77 & 6 & 3 & 14 & $-3$ \\
            & Sonnet 4.6 & DRAGO & 100 & 0 & 0 & 0 & $0$ \\
            & Sonnet 4.6 & LLM-diag. & 97 & 3 & 0 & 0 & $-3$ \\
\midrule
floor-tile & Haiku 4.5 & DRAGO & 21 & 1 & 0 & 78 & $-1$ \\
           & Haiku 4.5 & LLM-diag. & 18 & 4 & 2 & 76 & $-2$ \\
           & Sonnet 4.6 & DRAGO & 40 & 0 & 0 & 60 & $0$ \\
           & Sonnet 4.6 & LLM-diag. & 29 & 11 & 1 & 59 & $-10$ \\
\midrule
gripper & Haiku 4.5 & DRAGO & 69 & 0 & 1 & 30 & $+1$ \\
        & Haiku 4.5 & LLM-diag. & 67 & 2 & 9 & 22 & $+7$ \\
        & Sonnet 4.6 & DRAGO & 99 & 0 & 1 & 0 & $+1$ \\
        & Sonnet 4.6 & LLM-diag. & 82 & 17 & 1 & 0 & $-16$ \\
\bottomrule
\end{tabular}
\caption{Planetarium correctness transitions by subdomain for the hosted
models ($100$ instances per subdomain, single paired run).  The large
floor-tile W$\rightarrow$W mass contains the $58$ input--reference-inconsistent
instances (Appendix~\ref{app:benchmark-details}).  The baseline's only
positive cell (Haiku gripper, $+7$) coexists with regressions on every other
subdomain, most severely Sonnet gripper ($17$ breaks against $1$ fix).}
\label{tab:anthropic-transitions-pddl}
\end{table*}

\paragraph{Open-source correctness transitions.}
\label{app:opensource-transitions}
Table~\ref{tab:opensource-transitions} reports the correctness-transition breakdown of Figure~\ref{fig:fol-transitions} for the open-source track.  Each row decomposes the pooled run-example pairs of one benchmark/model/method cell into the four initial$\rightarrow$final buckets: R$\rightarrow$R (correct protected), R$\rightarrow$W (correct broken), W$\rightarrow$R (wrong fixed), and W$\rightarrow$W (still wrong); Net $=$ W$\rightarrow$R $-$ R$\rightarrow$W is the number of pairs the repair loop turns net-correct.  FOL transitions are measured from each method's \emph{own} initial formalization; Planetarium transitions use the paired v2 protocol, in which both repair arms start from the same saved zero-shot artifacts of each seed.  CSP is omitted because the CSP runners record only final correctness at the item level.

The asymmetry that drives the accuracy gap sits in the R$\rightarrow$W column.  DRAGO's break rate never exceeds $2.1\%$ of pooled pairs (FOLIO/Qwen3-Coder) and stays below $0.7\%$ on Planetarium for both models, whereas the LLM-diagnostic baseline breaks between $3.3\%$ and $14.0\%$ of the pairs it starts from.  On ProverQA-hard and Planetarium the baseline is net negative for both models: its ungated self-review destroys more correct formalizations than it repairs.  DRAGO is net positive in every cell, and not because it fixes more --- on FOLIO/Qwen3-Coder the baseline fixes more pairs ($135$ vs.\ $61$), and on Planetarium/Gemma it fixes $123$ against DRAGO's $80$ --- but because it does not break already-correct artifacts: on Planetarium the baseline pays for those fixes with $126$ (Gemma) and $85$ (Qwen) regressions, against DRAGO's $6$ and $5$.  The Planetarium W$\rightarrow$W mass includes the $58$ input--reference-inconsistent floor-tile instances of each seed (Appendix~\ref{app:benchmark-details}), which no repair method can recover.

\begin{table*}[t]
\centering
\setlength{\tabcolsep}{4.0pt}
\resizebox{\textwidth}{!}{%
\begin{tabular}{@{}lllrrrrrrr@{}}
\toprule
Benchmark & Model & Method & Seeds & $N$ &
R$\rightarrow$R & R$\rightarrow$W & W$\rightarrow$R & W$\rightarrow$W & Net \\
\midrule
FOLIO & Gemma 4 31B & DRAGO & 5 & 1010 & 851 & 8 & 34 & 117 & $+26$ \\
      & Gemma 4 31B & LLM-diag. & 5 & 1010 & 812 & 47 & 55 & 96 & $+8$ \\
      & Qwen3-Coder 30B & DRAGO & 5 & 1010 & 594 & 21 & 61 & 334 & $+40$ \\
      & Qwen3-Coder 30B & LLM-diag. & 5 & 1010 & 508 & 107 & 135 & 260 & $+28$ \\
\midrule
ProverQA-hard & Gemma 4 31B & DRAGO & 3 & 1500 & 1125 & 2 & 130 & 243 & $+128$ \\
              & Gemma 4 31B & LLM-diag. & 3 & 1500 & 1078 & 49 & 35 & 338 & $-14$ \\
              & Qwen3-Coder 30B & DRAGO & 3 & 1500 & 904 & 8 & 128 & 460 & $+120$ \\
              & Qwen3-Coder 30B & LLM-diag. & 3 & 1500 & 778 & 134 & 119 & 469 & $-15$ \\
\midrule
Planetarium & Gemma 4 31B & DRAGO & 3 & 900 & 545 & 6 & 80 & 269 & $+74$ \\
            & Gemma 4 31B & LLM-diag. & 3 & 900 & 425 & 126 & 123 & 226 & $-3$ \\
            & Qwen3-Coder 30B & DRAGO & 3 & 900 & 132 & 5 & 15 & 748 & $+10$ \\
            & Qwen3-Coder 30B & LLM-diag. & 3 & 900 & 52 & 85 & 20 & 743 & $-65$ \\
\bottomrule
\end{tabular}%
}
\caption{Open-source correctness transitions from each method's own initial
formalization to its final answer.  Counts are pooled over the seeds of each
cell, so $N$ is benchmark size times seeds; R denotes a correct artifact and W
an incorrect or unscored artifact.  Net $=$ W$\rightarrow$R $-$ R$\rightarrow$W.
CSP is omitted because the CSP runners do not record initial correctness.}
\label{tab:opensource-transitions}
\end{table*}

\subsection{Component Ablations}
\label{app:ablations}

\paragraph{Protocol.}
The ablation suite factorizes DRAGO's two mechanisms (the feedback source
that aims each rewrite and the acceptance predicate that decides whether to
keep it), holding everything else fixed.  All arms run at the one-round
budget of Table~\ref{tab:main-results} and are \emph{paired-start}: within
each configuration, every arm starts from the same saved initial artifacts
as the main-table runs, so differences between arms are attributable to the
repair stage alone.  Hosted-model cells are single paired runs on the
primary sweep of each configuration; open-model
cells use the same seeded runs as the main table.  Three ablation arms are
compared against the zero-shot, LLM-diagnostic, and DRAGO reference columns
of Table~\ref{tab:main-results}:
\begin{itemize}
  \item \textbf{DRAGO$-$A} (no acceptance): identical to DRAGO (same
  diagnostic operator, triggering condition, and repair prompts) but every
  proposed rewrite is retained unconditionally.
  \item \textbf{LLM-d$+$A} (gated self-review): identical to the
  LLM-diagnostic baseline, but each proposed rewrite is filtered through the
  same reference-free acceptance predicate that DRAGO uses in that
  formalism; rejected rewrites restore the previous artifact.
  \item \textbf{Solver-error-only}: repairs only artifacts that fail to
  parse, execute, or solve, using ordinary solver-error feedback; runnable
  artifacts are returned unchanged.  This arm isolates the share of the
  gains available outside the silent region.
\end{itemize}
All three arms cover the $20$ model--benchmark configurations of
Table~\ref{tab:main-results}.

\begin{table*}[t]
\centering
\setlength{\tabcolsep}{4.5pt}
\begin{tabular}{@{}llcccccc@{}}
\toprule
Benchmark & Model & Zero & Solv-err & LLM-d & LLM-d$+$A & DRAGO$-$A & DRAGO \\
\midrule
FOLIO & Haiku 4.5 & $75.0$ & $76.7$ & $81.5$ & $80.2$ & $79.2$ & $\mathbf{85.5}$ \\
      & Sonnet 4.6 & $85.0$ & $86.1$ & $81.2$ & $81.2$ & $86.1$ & $\mathbf{90.1}$ \\
      & Qwen3-Coder 30B & $60.9$ & $60.4$ & $63.7$ & $\mathbf{65.8}$ & $60.9$ & $64.8$ \\
      & Gemma 4 31B & $85.0$ & $85.2$ & $85.8$ & $84.2$ & $85.2$ & $\mathbf{87.6}$ \\
\midrule
ProverQA-hard & Haiku 4.5 & $77.1$ & $77.6$ & $74.4$ & $71.4$ & $77.0$ & $\mathbf{89.4}$ \\
      & Sonnet 4.6 & $77.2$ & $77.0$ & $74.1$ & $72.6$ & $83.0$ & $\mathbf{90.3}$ \\
      & Qwen3-Coder 30B & $60.8$ & $61.6$ & $59.8$ & $57.0$ & $63.0$ & $\mathbf{68.8}$ \\
      & Gemma 4 31B & $75.1$ & $75.2$ & $74.2$ & $75.6$ & $\mathbf{84.6}$ & $83.7$ \\
\midrule
Text2Zinc & Haiku 4.5 & $17.3$ & $34.5$ & $30.9$ & $32.7$ & $39.1$ & $\mathbf{41.8}$ \\
      & Sonnet 4.6 & $40.9$ & $49.1$ & $50.9$ & $53.6$ & $54.5$ & $\mathbf{56.4}$ \\
      & Qwen3-Coder 30B & $12.0$ & $17.3$ & $22.2$ & $\mathbf{26.4}$ & $23.1$ & $19.8$ \\
      & Gemma 4 31B & $20.4$ & $35.5$ & $38.7$ & $42.7$ & $52.7$ & $\mathbf{56.5}$ \\
\midrule
CP-Bench & Haiku 4.5 & $36.5$ & $47.6$ & $60.3$ & $57.1$ & $60.3$ & $\mathbf{61.9}$ \\
      & Sonnet 4.6 & $61.9$ & $71.4$ & $71.4$ & $\mathbf{79.4}$ & $\mathbf{79.4}$ & $\mathbf{79.4}$ \\
      & Qwen3-Coder 30B & $20.3$ & $23.8$ & $25.1$ & $22.2$ & $39.7$ & $\mathbf{42.9}$ \\
      & Gemma 4 31B & $43.9$ & $60.3$ & $59.3$ & $63.5$ & $66.7$ & $\mathbf{70.9}$ \\
\midrule
Planetarium & Haiku 4.5 & $58.0$ & $58.0$ & $58.7$ & $\mathbf{60.3}$ & $58.3$ & $\mathbf{60.3}$ \\
      & Sonnet 4.6 & $79.7$ & $79.7$ & $70.0$ & $75.0$ & $79.7$ & $\mathbf{80.0}$ \\
      & Qwen3-Coder 30B & $15.2$ & $15.7$ & $8.0$ & $8.0$ & $15.7$ & $\mathbf{16.3}$ \\
      & Gemma 4 31B & $61.2$ & $61.2$ & $60.9$ & $64.3$ & $68.7$ & $\mathbf{69.4}$ \\
\bottomrule
\end{tabular}
\caption{Component-ablation accuracy behind the
formalism-level aggregation of Figure~\ref{tab:ablation-summary}.  All
repair arms use the one-round budget and start from the same initial
artifacts. \textbf{Bold} marks the best
arm per row.}
\label{tab:ablation-full}
\end{table*}

\paragraph{Reading.}
Table~\ref{tab:ablation-full} adds three details to the main-text analysis
of Section~\ref{sec:ablations}.  First, the acceptance predicate's FOL
contribution is concentrated exactly where repairs can flip the final
label: on ProverQA-hard/Haiku, DRAGO$-$A nets zero against zero-shot
($77.0$ vs.\ $77.1$) while gated DRAGO gains $12.3$ points, so the ungated
loop fixes and breaks in roughly equal measure and the gate converts that
churn into gains.  Second, the configurations where an ablated arm beats
full DRAGO are informative rather than anomalous.  On Text2Zinc/Qwen every
ungated arm beats gated DRAGO, consistent with the main-text hypothesis
that these zero-shot artifacts are too poor for targeted repair; and on
ProverQA-hard/Gemma the ungated arm edges the gated one by $0.9$ points,
within the run-to-run noise of that cell.  In the remaining CSP
configurations the gate is neutral or helps: it ties on CP-Bench/Sonnet
and adds $1.6$--$4.2$ points elsewhere.  Third, gated self-review
(LLM-d$+$A) matches DRAGO only on Planetarium/Haiku and CP-Bench/Sonnet,
exceeds it only in the two Qwen cells above (FOLIO $+1.0$, Text2Zinc
$+6.6$), and loses by $2.8$--$20.7$ points in the remaining sixteen
configurations; gating tips self-review above zero-shot only in the two
Gemma cells where plain self-review was already within a point of neutral
(ProverQA-hard $+0.5$, Planetarium $+3.1$).  Where self-review is
destructive, gating at best mitigates: on Planetarium it halves Sonnet's
damage ($-9.7$ to $-4.7$) without turning it positive.  Acceptance alone
therefore reproduces DRAGO in no formalism across models.

\paragraph{Sampling baseline and oracle.}
The sampling baseline of Table~\ref{tab:sampling-summary} asks whether
DRAGO's gains could instead be obtained by resampling zero-shot generation
at the same call budget and selecting among the samples without a
reference.  For each hosted-model configuration it draws $k{=}3$ zero-shot
formalizations per instance and applies a selection rule fixed a priori,
which sees only the samples and their executor outcomes, never the
reference or the benchmark score:
\begin{itemize}
  \item \textbf{FOL}: majority vote over parseable predicted labels; a
  sample abstains if its label is unparseable or outside
  \{True, False, Uncertain\}; ties select the earliest run in the subset,
  and items where no sample produces a valid label are scored wrong.
  \item \textbf{CSP}: for optimization instances, each solved sample
  reports its own objective value, and a value reported by at least two
  samples selects the earliest sample reporting it, otherwise the earliest
  solved sample; for satisfaction instances, the earliest solved sample;
  if no sample solves, the earliest sample.
  \item \textbf{PDDL}: parseable candidates are clustered with the official
  evaluator's graph-equivalence procedure (applied only between candidates,
  never against the reference; an equivalence-check timeout counts as not equivalent);
  the earliest member of the largest cluster is selected, with ties
  resolving to the cluster containing the earliest candidate; if no
  candidate parses, the earliest sample is returned.
\end{itemize}
The \emph{oracle-of-3} scores an item correct if \emph{any} of the three
samples is correct under the official scorer; it \textbf{upper-bounds} every
conceivable selection strategy at this budget, including selection by a
reference-aware judge.  Because sampling was run only for
the hosted models, the Zero and DRAGO columns of
Table~\ref{tab:sampling-summary} average the primary-run hosted cells
behind Table~\ref{tab:main-results} and therefore differ from the
all-model aggregation of Figure~\ref{tab:ablation-summary}.

Table~\ref{tab:sampling-full} reports the ten hosted configurations.  The
disagreement structure of the three samples explains the oracle's low
ceiling: zero-shot errors are dominated by items on which every sample
encodes the same wrong semantics or fails to solve at all. On
ProverQA-hard/Sonnet, $7$ of $500$ items show any cross-sample
disagreement, and on Planetarium/Sonnet, $1$ of $300$, which is exactly
the error mass that selection cannot touch and diagnosis-guided repair
can.  Sampling helps most on CSP ($+0.9$ to $+11.1$ points over
zero-shot), where its gains coincide with the execution-failure region
also recovered by the solver-error-only arm; even there the oracle stays
$11.1$--$18.2$ points below DRAGO.

\paragraph{Sampling provenance.}
All samples reuse the repeated zero-shot runs behind
Table~\ref{tab:main-results};
FOLIO $k{=}3$ averages all $\binom{5}{3}=10$ run subsets, and accuracy
is flat beyond $k{=}3$.  Runs are aligned by problem name, except PDDL,
aligned by row index because eight problem names recur across
abstraction levels.

\begin{table}[t]
\centering
\setlength{\tabcolsep}{4.0pt}
\begin{tabular}{@{}lcccc@{}}
\toprule
Model & Zero & Sample-3 & Oracle-3 & DRAGO \\
\midrule
\multicolumn{5}{@{}l}{\emph{FOLIO}} \\
Haiku 4.5 & $75.0$ & $78.6$ & $80.2$ & $\mathbf{85.5}$ \\
Sonnet 4.6 & $85.0$ & $86.1$ & $87.6$ & $\mathbf{90.1}$ \\
\midrule
\multicolumn{5}{@{}l}{\emph{ProverQA-hard}} \\
Haiku 4.5 & $77.1$ & $78.2$ & $79.2$ & $\mathbf{89.4}$ \\
Sonnet 4.6 & $77.2$ & $77.2$ & $78.0$ & $\mathbf{90.3}$ \\
\midrule
\multicolumn{5}{@{}l}{\emph{Text2Zinc}} \\
Haiku 4.5 & $17.3$ & $23.6$ & $23.6$ & $\mathbf{41.8}$ \\
Sonnet 4.6 & $40.9$ & $41.8$ & $42.7$ & $\mathbf{56.4}$ \\
\midrule
\multicolumn{5}{@{}l}{\emph{CP-Bench}} \\
Haiku 4.5 & $36.5$ & $47.6$ & $50.8$ & $\mathbf{61.9}$ \\
Sonnet 4.6 & $61.9$ & $63.5$ & $66.7$ & $\mathbf{79.4}$ \\
\midrule
\multicolumn{5}{@{}l}{\emph{Planetarium}} \\
Haiku 4.5 & $58.0$ & $59.3$ & $63.7$ & $\mathbf{60.3}$ \\
Sonnet 4.6 & $79.7$ & $\mathbf{80.0}$ & $80.0$ & $\mathbf{80.0}$ \\
\bottomrule
\end{tabular}
\caption{Per-configuration accuracy of the sampling baseline behind
Table~\ref{tab:sampling-summary}.  Zero and DRAGO report the primary
paired run behind Table~\ref{tab:main-results} (FOL cells are means over
the repeated runs); Sample-3 is the reference-free selector and Oracle-3
the any-correct ceiling defined above.  \textbf{Bold} marks the best
implementable method per row.  Planetarium/Haiku is the one configuration
where the oracle exceeds every implementable arm; realizing it would
require the reference-based equivalence judgment of the benchmark scorer
itself.}
\label{tab:sampling-full}
\end{table}

\paragraph{Draft-and-Prune baseline.}
Draft-and-Prune~\cite{ni2026draftandpruneimprovingreliabilityautoformalization}
runs each reasoning path in three steps: \emph{draft} a natural-language
plan of the premises' logical structure (sampled at temperature $1.0$),
\emph{code} the plan into one symbolic program (decoded greedily), and
\emph{prune} by execution, with a bounded syntax-repair loop of up to
three attempts before a path is discarded; surviving paths are
majority-voted over $k$ paths.  We evaluate the single-path configuration
$k{=}1$, whose budget (two generation calls per item plus bounded syntax
repair) is comparable to DRAGO's; the $k{=}5$ ensemble of the original
paper spends $3$--$5\times$ the inference of every method in our
comparison.  At
$k{=}1$ a path that fails to run is scored wrong with no sibling path to
fall back on; this cost of the compute-matched setting affects $26/202$
FOLIO and $83/500$ ProverQA-hard items for Qwen and $8$ and $34$ for
Gemma.  The released implementation does not include an executable FOL
path (its solver map names Prover9, but no Prover9 executor, FOLIO
prompts, or ProverQA support is present), so we complete it: the symbolic program
is SMT-LIB2 executed by Z3 under the same output contract, evaluator,
gold predicate signature, and model clients as every other method in this
paper, so the comparison isolates the strategy (planning, pruning,
ensembling) rather than the solver backend.  Each Draft-and-Prune cell is
a single run and the hosted models were not
evaluated for this baseline due to compute budget.  Reading
Table~\ref{tab:dnp} as deltas over zero-shot: on ProverQA-hard the plan
step adds $+4.6$ (Qwen) and $+0.3$ (Gemma) points where DRAGO adds
$+8.0$ and $+8.6$; on FOLIO/Qwen planning wins by $5.5$ points at the
weakest zero-shot base ($60.9$), and it adds essentially nothing where
the base model already reads the premises correctly.

\paragraph{CP-SynC baseline.}
CP-SynC~\cite{song2026cpsyncmultiagentzeroshotconstraint} spends its budget on parallel synthesis and
checking rather than on repair: $K$ modeling agents each draft a MiniZinc
model, $K$ validation agents each synthesize a Python semantic checker,
and every candidate runs a staged cascade (G1 syntax
(\texttt{--model-check-only}), G2 solver status, G3 output format, G4
checker majority) with targeted repair ($r{=}4$ per candidate) and up to
$R{=}1$ restart that rewrites the problem description; $K$ selection
agents then vote on the survivors.  We re-implement the system, as no
executable release was available, and run it on the same MiniZinc
backend, model clients, and scorer as every other method in this paper.
Diversity comes from temperature sampling at CP-SynC's setting
($\tau{=}0.7$), with the sampling seed advanced per call so the $K$
same-prompt draws differ while the run stays reproducible. Each CP-SynC cell is a single run and its
diversity is induced by temperature sampling, so
run-to-run variance is real but unmeasured, and the two closest calls
against DRAGO (Text2Zinc/Qwen at $k{=}1$ and CP-Bench/Qwen at
$k{=}3$) lie within DRAGO's own seed-to-seed variability, so these gaps should not be treated as statistically meaningful.
The hosted models were not evaluated for this baseline due to compute
budget.

Table~\ref{tab:cpsync-full} gives the full CSP comparison, adding the
$k{=}3$ arm that Table~\ref{tab:dnp} defers here and the LLM-diagnostic
baseline for reference.  Inference budget is measured as mean LLM calls
per item over these runs.  Neither CP-SynC arm is budget-comparable to
DRAGO: $k{=}1$ already spends $3.8\times$ DRAGO's calls and $k{=}3$
spends $10.1\times$.  DRAGO leads three of the four cells at both arms,
and its $1.6$-point margin over $k{=}3$ on CP-Bench/Qwen sits inside its
own $\pm2.7$ seed spread---a tie bought at ten times the calls.  The
$k{=}1\rightarrow k{=}3$ gain is coverage rather than sharper modeling:
solved counts rise in every cell (CP-Bench/Qwen $30\rightarrow42$,
CP-Bench/Gemma $34\rightarrow44$, Text2Zinc/Qwen $54\rightarrow62$,
Text2Zinc/Gemma $62\rightarrow79$) while accuracy conditional on being
solved is roughly flat,  so the extra trajectories mostly convert unsolved
items into solved ones; where conditional accuracy is higher at $k{=}3$,
this is a selection effect rather than evidence of more
accurate modeling.  At $k{=}3$ CP-SynC beats
LLM-diagnostic repair in every cell, by $+4.2$ to $+16.2$ points.

\begin{table}[t]
\centering
\setlength{\tabcolsep}{2.5pt}
\resizebox{\columnwidth}{!}{%
\begin{tabular}{@{}lccccc@{}}
\toprule
& & & \multicolumn{2}{c}{CP-SynC} & \\
\cmidrule(lr){4-5}
Model & Zero & LLM-d & $k{=}1$ & $k{=}3$ & DRAGO \\
\midrule
\multicolumn{6}{@{}l}{\emph{CP-Bench}} \\
Qwen3-Coder 30B & $20.3$ & $25.1$ & $27.0$ & $41.3$ & $\mathbf{42.9}$ \\
Gemma 4 31B & $43.9$ & $59.3$ & $47.6$ & $63.5$ & $\mathbf{70.9}$ \\
\midrule
\multicolumn{6}{@{}l}{\emph{Text2Zinc}} \\
Qwen3-Coder 30B & $12.0$ & $22.2$ & $20.0$ & $\mathbf{30.0}$ & $19.8$ \\
Gemma 4 31B & $20.4$ & $38.7$ & $41.8$ & $48.2$ & $\mathbf{56.5}$ \\
\midrule
Calls/item & --- & $3.57$ & $10.01$ & $26.66$ & $2.65$ \\
\bottomrule
\end{tabular}%
}
\caption{Full CP-SynC comparison on CSP, at both the single-trajectory
arm ($k{=}1$) reported in Table~\ref{tab:dnp} and the paper-default
$k{=}3$.  Zero, LLM-d and DRAGO repeat the multi-run means of
Table~\ref{tab:main-results}; each CP-SynC cell is a single run and
therefore a point estimate.  \textbf{Bold} marks the best method per row;
the CP-Bench/Qwen ($42.9\pm2.7$ vs.\ $41.3$) and Text2Zinc/Qwen
($19.8\pm1.8$ vs.\ $20.0$ at $k{=}1$) margins are within run-to-run
noise.  The final row reports the measured inference budget as mean LLM
calls per item, against which every accuracy column should be read.}
\label{tab:cpsync-full}
\end{table}

\subsection{Failure Analysis}
\label{app:failure-analysis}

\begin{figure}[t]
    \centering
    \includegraphics[width=1\linewidth]{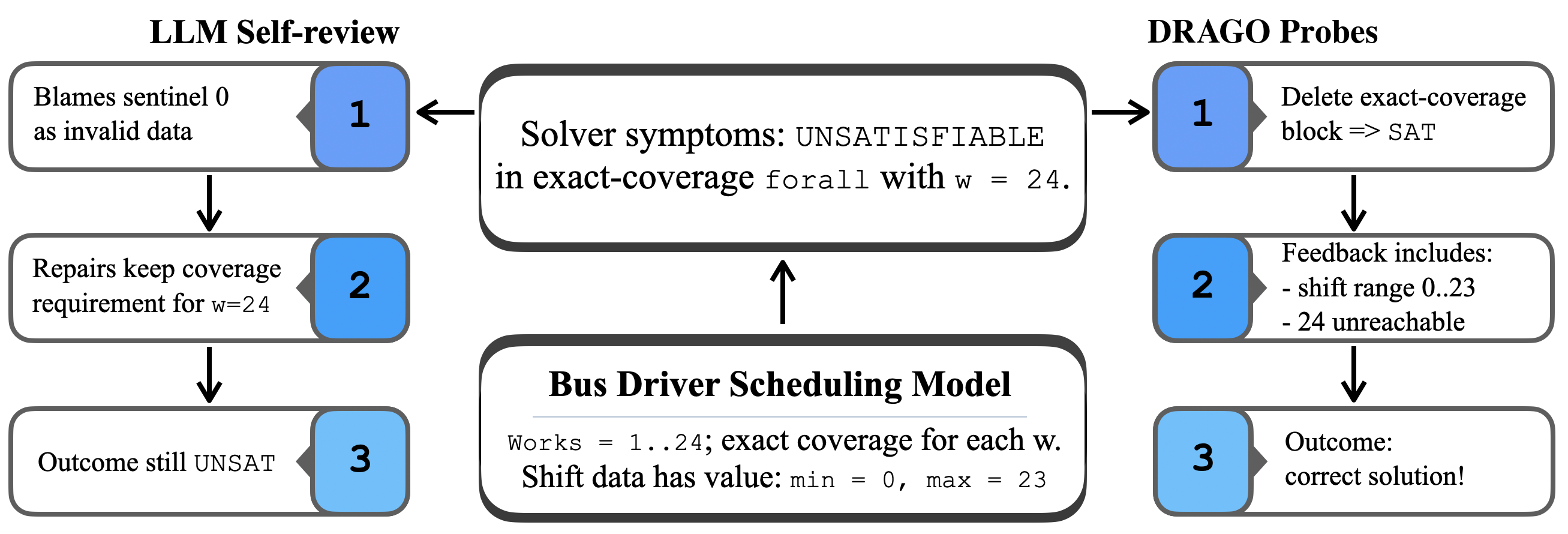}
    \caption{Case study on an UNSAT CSP instance (Bus Driver Scheduling), contrasting the diagnoses of LLM self-review and DRAGO.}
    \label{fig:csp-unsat-feedback}
\end{figure}

\paragraph{MiniZinc UNSAT case studies.}
Aggregate results do not show \emph{what} information a repair model needs.  A representative CSP failure, Bus Driver Scheduling, puts three feedback signals of very different value on the same fault: the generated model requires exact coverage for work items \texttt{1..24}, but the data's identifiers span \texttt{0..23} with \texttt{0} a sentinel, so no shift covers item \texttt{24}.  The raw solver trace reports \texttt{UNSATISFIABLE} at \texttt{w = 24}: a symptom, not an edit.  LLM self-review misdiagnoses the sentinel \texttt{0} as invalid data, and its repair stays \texttt{UNSAT}.  DRAGO's deletion probes find that removing the exact-coverage block restores satisfiability and report the data range \texttt{0..23}; the accepted repair restricts coverage to reachable work items and the solver returns a correct solution (Figure~\ref{fig:csp-unsat-feedback}).  Grounded diagnosis thus turns a failure status into a concrete repair target.

A second UNSAT case, Wolf-Goat-Cabbage, illustrates the same grounded feedback on a more intuitive transition system.  The generated model was syntactically valid but unsatisfiable: with \texttt{stage = 8}, the boat-alternation constraint forces eight crossings that return the boat to the starting shore, contradicting the required final boat position.  The raw solver trace reports only \texttt{UNSATISFIABLE}, whereas DRAGO's deletion probes single out the boat-alternation block as the unique tested deletion that restores satisfiability, pointing the repair model to the horizon interpretation rather than to unrelated movement details.  The accepted repair re-indexes the stage horizon (final state at \texttt{stage-1}, transitions over \texttt{0..stage-2}), after which the solver returns a correct solution.

\paragraph{Execution-blind under-constraint: a FOL example.}
As stated in the Limitations section,
DRAGO's main blind spot is under-constraint that executes cleanly; we
present an example from FOLIO (reference label True).  The premises
state that there are six types of wild turkey, that Tom is not any of
the first five, and that Tom is a wild turkey; deriving the conclusion
(Tom is an ocellated wild turkey) requires the closure axiom
$\forall x.\,\mathit{WildTurkey}(x) \rightarrow
\mathit{T}_1(x) \lor \cdots \lor \mathit{Ocellated}(x)$.
DRAGO's formalization instead asserted only the six subtype implications
$\mathit{T}_i(x) \rightarrow \mathit{WildTurkey}(x)$ (each faithful to
a sentence, jointly too weak to force the conclusion) so Z3 returned
\emph{undetermined} and the pipeline answered Uncertain.  The script
parses, executes, and yields a legitimate label; no contract check,
solver probe, or alignment check fires, because the error is a missing
axiom rather than a wrong one.  The failure is direction, not coverage:
the failing script contains \emph{more} assertions than the
Draft-and-Prune script that succeeds on the same item ($12$ vs.\ $6$).
At corpus scale the effect is label-asymmetric: over the undetermined
verdicts of a representative Qwen DRAGO run, every Uncertain-labeled
item is scored correct by construction ($62/62$ on FOLIO, $146/146$ on
ProverQA-hard), while every True- or False-labeled item is scored wrong
($37$ and $73$ items): strong Uncertain recall, leaking True/False
recall.  Draft-and-Prune sidesteps the failure by writing the closure
axiom in natural language before any code, after which Z3 proves
entailment.  Detecting under-constraint reference-free would require
reasoning about what the premises should force, which our current FOL finding classes
do not attempt.

\subsection{Licenses}
\label{app:licenses}

All benchmarks are used under their published terms, for research
evaluation only; Table~\ref{tab:licenses} lists their licenses.  The only
prior artifact we redistribute is Planetarium: our corrected floor-tile
split is a derivative work, which CC BY 4.0 permits, and it is released
under the same license with attribution
to~\citet{zuo-etal-2025-planetarium} and a description of the changes.
Our code is released under Apache 2.0.

\begin{table}[h]
\centering
\small
\begin{tabular}{@{}ll@{}}
\toprule
Artifact & License \\
\midrule
FOLIO~\cite{han2024folionaturallanguagereasoning} & CC BY-SA 4.0 \\
ProverQA~\cite{qi2025large} & none stated \\
Text2Zinc~\cite{singirikonda2025text2zinccrossdomaindatasetmodeling} & CC BY 4.0 \\
CP-Bench~\cite{michailidis2026dcpbenchopenevaluatingllmsconstraint} & Apache 2.0 \\
Planetarium~\cite{zuo-etal-2025-planetarium} & CC BY 4.0 \\
\midrule
Corrected floor-tile split (ours) & CC BY 4.0 \\
DRAGO code (ours) & Apache 2.0 \\
\bottomrule
\end{tabular}
\caption{Licenses of the benchmarks used and of the artifacts we release.
ProverQA is publicly released by its authors without a stated license.}
\label{tab:licenses}
\end{table}